%% file: main.tex
\documentclass[journal]{vgtc}              

\onlineid{1461}

\vgtccategory{Research}

\vgtcpapertype{Data Transformations}

\title{A Practical Solver for Scalar Data Topological Simplification}

\author{%
  Mohamed Kissi,
  Mathieu Pont,
  Joshua A. Levine,
  Julien Tierny
}

\authorfooter{
  \item Mohamed Kissi, Mathieu Pont, and Julien Tierny are with the CNRS and
Sorbonne University.
  	E-mail: \{firstname.lastname\}@sorbonne-universite.fr
  \item
    Joshua A. Levine is with the University of Arizona.\\
  	E-mail: josh@cs.arizona.edu.
}

\abstract{\input{abstract.tex}}


\teaser{
  \centering
  \includegraphics[width=\linewidth]{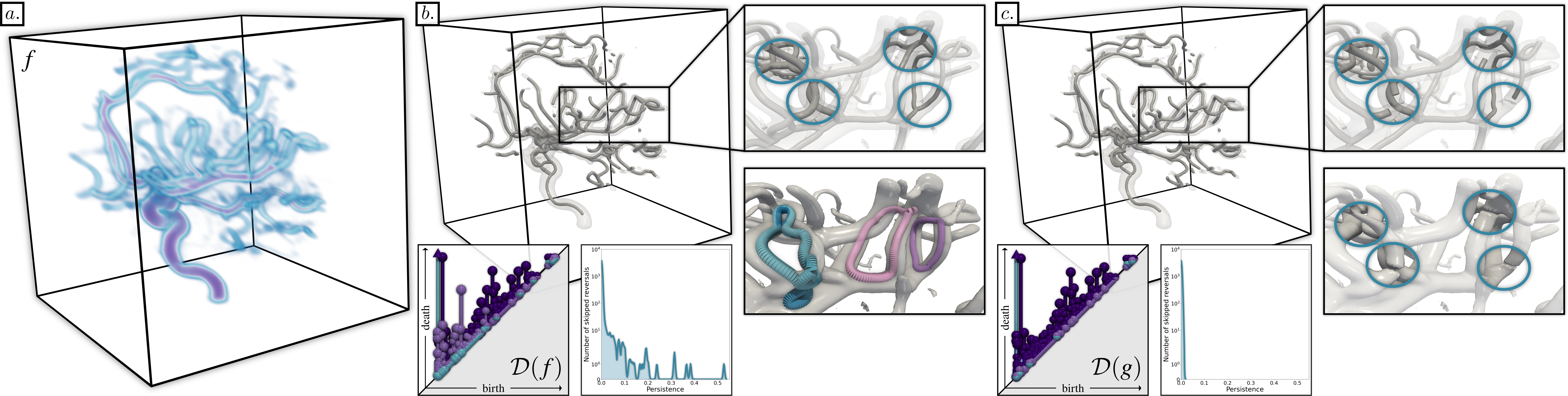}
  \caption{Given an
  acquired scalar field $f$ of a network of
arteries $(a)$, the core structure of the blood vessels can be extracted (grey
filaments, $(b)$)
with upward discrete integral lines, started at $2$-saddles above $0.1$
(isovalue
representing the geometry of the vessels,
transparent isosurface). However, as shown in the persistence diagram
$\diagram(f)$, $f$ contains many saddle-pairs (light purple bars),
corresponding to persistent $1$-dimensional generators \cite{guillou_tvcg23,
iuricich21} (curves, colored by persistence, bottom zoom, $(b)$),
yielding incorrect loops in the filament structures (top zoom). In
this example, standard techniques for gradient field simplification (i.e.,
saddle connector reversal) cannot simplify these spurious loops while
maintaining a valid gradient $(b)$, as shown in the bottom histogram (number of
\emph{skipped} reversals as a function of persistence).
Our approach efficiently generates a function $g$
which is close to
$f$ and
which optimizes saddle pair cancellation
while maintaining the other features
($\diagram(g)$). This enables the direct
visualization and analysis of the
simplified data $(c)$, where
isosurface handles have been cut (bottom zoom) and
most spurious filament loops
have
been
simplified (top zoom).}
  \label{fig_teaser}
}

\graphicspath{{figs/}{figures/}{pictures/}{images/}{./}} 

\usepackage{booktabs}                  
\usepackage{lipsum}                    
\usepackage{amssymb}

\usepackage{mathptmx}                  
\usepackage{calrsfs}
\usepackage{algorithm}
\usepackage{algorithmicx}
\usepackage{algpseudocode}

\usepackage{multirow}
\usepackage{hhline}
\usepackage{colortbl}
\usepackage{adjustbox}
\usepackage{enumitem}

\algdef{SE}[DOWHILE]{Do}{doWhile}{\algorithmicdo}[1]{\algorithmicwhile\ #1}%

\DeclareMathAlphabet{\pazocal}{OMS}{zplm}{m}{n}
\SetMathAlphabet\pazocal{bold}{OMS}{zplm}{bx}{n}

\begin{document}


\newcommand{\revision}[1]{\textcolor{blue}{#1}}
\input{notations.tex}
\input{introduction.tex}

\input{relatedWork.tex}
\input{contributions.tex}

\input{background.tex}

\input{approach.tex}

\input{algorithm.tex}

\input{results.tex}

\input{conclusion.tex}

\vspace{-1ex}
\acknowledgments{
\small{
\vspace{-1ex}
\revision{This work is partially supported by the European Commission grant
ERC-2019-COG \emph{``TORI''} (ref. 863464, \url{https://erc-tori.github.io/}),
by the U.S. Department of Energy, Office of Science,
under Award Number(s)
DE-SC-0019039,
and by a joint graduate research fellowship (ref. 320650) funded by the CNRS
and the University of Arizona.}}
\normalsize
\vspace{-1ex}
}


\bibliographystyle{abbrv-doi-hyperref}

\bibliography{main}

 \clearpage
 \appendix
 \section*{Appendix}
 \includegraphics{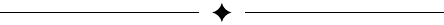}
 \input{appendixBody}


\end{document}

%% file: abstract.tex
This paper presents a practical approach for the optimization of
topological simplification, a central pre-processing step for the analysis and
visualization of scalar data. Given an input scalar field $f$ and
a set of \emph{``signal''} persistence pairs to maintain,
our approaches produces an output field $g$ that is close to $f$ and
which optimizes
\emph{(i)} the cancellation of \emph{``non-signal''} pairs,
while \emph{(ii)}
preserving the \emph{``signal''} pairs.
In contrast to pre-existing simplification \revision{algorithms,}
our \revision{approach}
is not restricted to persistence pairs involving extrema
and can thus address a larger class of topological features, in particular
saddle pairs in three-dimensional scalar data. Our approach leverages recent
generic persistence optimization frameworks and extends them with tailored
accelerations specific to the problem of topological simplification. Extensive
experiments report
substantial
accelerations
over
these
frameworks,
thereby
making
topological simplification
optimization
practical for real-life datasets.
Our
\revision{approach}
enables a direct visualization and
analysis of the topologically simplified data, e.g., via isosurfaces of
simplified topology (fewer components and handles).
We apply our
approach
to the extraction of
prominent filament structures in three-dimensional data.
Specifically, we show
that our pre-simplification of the data leads to practical improvements
over standard topological techniques
for removing filament loops.
We also show how our
\revision{approach}
can
be used to
repair genus defects in surface processing.
Finally, we
provide a C++ implementation
for reproducibility
purposes.

%% file: notations.tex
\renewcommand{\mathcal}[1]{\pazocal{#1}}

\newcommand{\surface}{S}

\newcommand{\signal}{\mathcal{S}}
\newcommand{\noise}{\mathcal{N}}
\newcommand{\simplex}{\sigma}
\newcommand{\domain}{\mathcal{K}}
\newcommand{\numberOfVertices}{n_v}
\newcommand{\numberOfSimplices}{n_\simplex}
\newcommand{\dataVector}{v}
\newcommand{\dataVectorSpace}{\mathcal{V}}
\newcommand{\filtration}{\mathcal{F}}
\newcommand{\persistenceMap}{\mathcal{P}}
\newcommand{\energy}{\mathcal{E}}
\newcommand{\loss}{\mathcal{L}}
\newcommand{\range}{\mathbb{R}}
\newcommand{\sublevelset}[1]{#1^{-1}_{-\infty}}
\newcommand{\superlevelset}[1]{#1^{-1}_{+\infty}}
\newcommand{\Star}{St}
\newcommand{\Link}{Lk}
\newcommand{\diagram}{\mathcal{D}}
\newcommand{\target}{\diagram_T}
\newcommand{\complexity}{\mathcal{O}}

\newcommand{\face}{\tau}
\newcommand{\lowerlink}{\Link^{-}}
\newcommand{\upperlink}{\Link^{+}}
\newcommand{\Index}{\mathcal{I}}
\newcommand{\offset}{o}
\newcommand{\Natural}{\mathbb{N}}
\newcommand{\criticalSet}{\mathcal{C}}

\newcommand{\wasserstein}[1]{\mathcal{W}_{#1}}
\newcommand{\projection}{\Delta}
\newcommand{\hierarchy}{\mathcal{H}}
\newcommand{\decimation}{D}
\newcommand{\xDimD}{L_x^\decimation}
\newcommand{\yDimD}{L_y^\decimation}
\newcommand{\zDimD}{L_z^\decimation}
\newcommand{\xDim}{L_x}
\newcommand{\yDim}{L_y}
\newcommand{\zDim}{L_z}
\newcommand{\Grid}{\mathcal{G}}
\newcommand{\GridD}{\mathcal{G}^\decimation}
\newcommand{\x}{\phantom{x}}
\newcommand{\Mod}{\;\mathrm{mod}\;}
\newcommand{\NN}{\mathbb{N}}
\newcommand{\forwardIntegralLine}{\mathcal{L}^+}
\newcommand{\backwardIntegralLine}{\mathcal{L}^-}
\newcommand{\triangulationOp}{\phi}
\newcommand{\decimationOp}{\Pi}
\newcommand{\isovalue}{w}
\newcommand{\persistence}{p}
\newcommand{\pointMetric}{d}
\newcommand{\diagramSet}{\mathcal{S}_\mathcal{D}}
\newcommand{\diagramSpace}{\mathbb{D}}
\newcommand{\jointree}{\mathcal{T}^-}
\newcommand{\splittree}{\mathcal{T}^+}
\newcommand{\mergetree}{\mathcal{T}}
\newcommand{\tree}{\mergetree}
\newcommand{\depth}{d}
\newcommand{\mergetreeSet}{\mathcal{S}_\mathcal{T}}
\newcommand{\branchset}{\mathcal{S}_\mathcal{B}}
\newcommand{\branchspace}{\mathbb{B}}
\newcommand{\mergetreeSpace}{\mathbb{T}}
\newcommand{\editdistance}{D_E}
\newcommand{\wassersteinTree}{W^{\mergetree}_2}
\newcommand{\distanceSequence}{d_S}
\newcommand{\branchtree}{\mathcal{B}}
\newcommand{\branchtreeSet}{\mathcal{S}_\mathcal{B}}
\newcommand{\branchtreeSpace}{\mathbb{B}}
\newcommand{\forest}{\mathcal{F}}
\newcommand{\sequenceSpace}{\mathbb{S}}
\newcommand{\forestMatrix}{\mathbb{F}}
\newcommand{\treeMatrix}{\mathbb{T}}
\newcommand{\normalizedLocation}{\mathcal{N}}
\newcommand{\normalizedWasserstein}{W^{\normalizedLocation}_2}
\newcommand{\geodesictree}{\mathcal{G}}
\newcommand{\dummyVector}{\mathcal{V}}
\newcommand{\geodesictreeVec}{g}
\newcommand{\geodesicAxis}{\mathcal{A}}
\newcommand{\directionVector}{\mathcal{V}}
\newcommand{\geodesicdiagram}{\mathcal{G}^{\diagram}}
\newcommand{\reconstructionError}{E_{L_2}}
\newcommand{\pcaBasis}{B_{\mathbb{R}^d}}
\renewcommand{\pcaBasis}{B}
\newcommand{\origin}{o_b}
\newcommand{\sizeEncoding}{n_e}
\newcommand{\sizeDecoding}{n_d}
\newcommand{\linearTransformation}{\psi}
\newcommand{\unitTransformation}{\Psi}
\renewcommand{\origin}{o}
\newcommand{\bdtOrigin}{\mathcal{O}}
\newcommand{\activation}{\sigma}
\newcommand{\validBDT}{\gamma}
\newcommand{\mtPgaBasis}{B_{\branchtreeSpace}}
\newcommand{\mtPgaError}{E_{\wassersteinTree}}
\newcommand{\frechetEnergy}{E_F}
\newcommand{\geodesicExtremity}{\mathcal{E}}
\newcommand{\vectorNotation}[1]{\protect\vv{#1}}
\renewcommand{\vectorNotation}[1]{#1}
\newcommand{\axisNotation}[1]{\protect\overleftrightarrow{#1}}
\newcommand{\individualEnergy}{E}
\newcommand{\ensembleSize}{N}
\newcommand{\numberBranchinBarycenter}{N_1}
\newcommand{\numberGeodesicSamples}{N_2}
\newcommand{\planarGridX}{N_x}
\newcommand{\planarGridY}{N_y}
\newcommand{\regularGrid}{G}
\newcommand{\distanceMatrix}{\mathbb{D}}
\newcommand{\maxDimensions}{{d_{max}}}
\newcommand{\projectionOperator}{\mathcal{P}}
\newcommand{\reconstructed}[1]{\widehat{#1}}
\newcommand{\gt}{>}
\newcommand{\lt}{<}
\newcommand{\branch}{b}
\newcommand{\nonLinearFunction}{\sigma}
\newcommand{\batchSequence}{S}
\newcommand{\homologyGroup}{\mathcal{H}}
\newcommand{\bettiNumber}{\beta}
\newcommand{\still}{\mathcal{S}}

\newcommand{\julien}[1]{\textcolor{red}{#1}}
\renewcommand{\julien}[1]{\textcolor{black}{#1}}
\newcommand{\mathieu}[1]{\textcolor{green}{#1}}
\renewcommand{\mathieu}[1]{\textcolor{green}{#1}}
\newcommand{\jules}[1]{\textcolor{orange}{#1}}
\renewcommand{\jules}[1]{\textcolor{black}{#1}}
\newcommand{\note}[1]{\textcolor{magenta}{#1}}
\newcommand{\cutout}[1]{\textcolor{blue}{#1}}
\renewcommand{\cutout}[1]{}

\renewcommand{\revision}[1]{\textcolor{black}{#1}}
\newcommand{\minor}[1]{\textcolor{blue}{#1}}

\newcommand{\discuss}[1]{\textcolor{black}{#1}}

\renewcommand{\figureautorefname}{Fig.}
\renewcommand{\sectionautorefname}{Sec.}
\renewcommand{\subsectionautorefname}{Sec.}
\renewcommand{\equationautorefname}{Eq.}
\renewcommand{\tableautorefname}{Tab.}
\newcommand{\algorithmautorefname}{Alg.}
\newcommand{\lineautorefname}{Alg.}

\newcommand{\todo}[1]{\textcolor{red}{TODO: #1}}

\newcommand{\mycaption}[1]{
\caption{#1}
}

\newcommand{\journal}[1]{\textcolor{blue}{#1}}
\renewcommand{\journal}[1]{\textcolor{black}{#1}}



%% file: introduction.tex
\firstsection{Introduction}

\maketitle

\label{sec_intro}

As acquisition devices and computational resources are getting more
sophisticated
and efficient,
modern datasets are growing in size.
Consequently,
the features of interest contained in these datasets gain in
\revision{geometric}
complexity, which challenge their interpretation and analysis.
This motivates the design of tools capable of robustly extracting the
structural patterns hidden in complex datasets.
This
task
is the purpose of Topological Data Analysis (TDA) \cite{edelsbrunner09,
zomorodianBook}, which provides a family of techniques for the generic, robust
and multi-scale extraction of structural features.
It has been successfully applied in a number of data analysis problems
\cite{heine16}, in various applications, including
turbulent combustion \cite{bremer_tvcg11, gyulassy_ev14},
material sciences \cite{gyulassy_vis15, soler_ldav19},
nuclear energy \cite{beiNuclear16},
fluid dynamics \cite{kasten_tvcg11, nauleau_ldav22},
bioimaging \cite{topoAngler, beiBrain18},
quantum chemistry \cite{harshChemistry, Malgorzata19, olejniczak_pccp23},
or astrophysics \cite{sousbie11, shivashankar2016felix}.
TDA provides a variety of \emph{topological data abstractions}, which
enable the extraction of specific types of features of interest. These
abstractions include
critical points \cite{banchoff70},
persistence diagrams \cite{edelsbrunner02, dipha, guillou_tvcg23},
merge \cite{CarrWSA16,
gueunet_ldav17, LukasczykWWWG24} and contour trees \cite{carr00,
gueunet_tpds19}, Reeb graphs \cite{biasotti08, pascucci07,
gueunet_egpgv19}, or Morse-Smale complexes
\cite{gyulassy_vis08, robins_pami11, ShivashankarN12, gyulassy_vis18, robin23}.
A central aspect of TDA is its ability to analyze data at multiple scales.
Thanks to various importance measure \cite{edelsbrunner02, carr04}, these
abstractions can be
iteratively simplified, to reveal the prominent structures in a dataset.

In practice, this topological simplification can be achieved in two fashions:
either by \emph{(i)} a post-process simplification of the abstractions, or by
\emph{(ii)} a
pre-process simplification of the data itself. While the post-process approach
\emph{(i)}
requires specific simplification mechanisms tailored to the abstraction at hand
\cite{carr04, pascucci07, gyulassy_phd08}, the pre-process strategy offers a
generic framework which is independent of the considered abstraction. This
generic aspect eases software design, as simplification needs to be implemented
only once \cite{ttk17, ttk19}. Pre-process simplification has also the
advantage of being reusable by multiple abstractions when these are combined
within a single data analysis pipeline (see \cite{ttkExamples} for real-life
examples). Also, pre-process
simplification enables the direct visualization of the simplified data itself
(e.g. with isosurfaces).
Finally, it is also compatible with further post-process simplification if
needed. For these reasons, we focus  on pre-process
simplification in this work.

Several
combinatorial approaches \cite{soille04, EdelsbrunnerMP06, AgarwalAY06,
attali09, BauerLW12, tierny_vis12, Lukasczyk_vis20}
have been proposed for the
pre-simplification of
persistence
pairs involving
extrema. However, no
efficient combinatorial algorithm has been proposed for the pre-simplification
of saddle pairs, hence preventing
a more advanced
simplification of 3D datasets.
\revision{In fact, as pointed out by Chambers et al. \cite{ChambersJLLTY18},
\emph{optimal simplification}  (i.e. finding a scalar field $g$
which is $\delta$-away from an input field $f$, such that $g$
has a minimum
number of critical points \cite{BauerLW12}) is a more general, hence more
difficult, version of the \emph{sublevel set simplification} problem, itself
being NP-hard in 3D \cite{AttaliBDGL13}.}
Then, there may not even exist polynomial time
algorithms for directly solving this problem.
This theoretical limitation
requires a shift in strategy. A recent alternative consists in
considering persistence optimization frameworks \cite{CarriereCGIKU21,
SolomonWB21,
Nigmetov22}, which optimize
the data in a \emph{best effort} manner,
given criteria expressed with persistence diagrams.
However, while one could leverage these
frameworks for  data pre-simplification (i.e., to
cancel
noisy features while preserving the
features of interest,  \emph{as much as possible}),
current frameworks
can
require \discuss{up to days of computation for regular grids of standard size}
(\autoref{sec_approach}),
making
them impractical for real-life datasets.

This paper addresses this issue and introduces a practical solver for the
optimization of the topological simplification of scalar data. Our approach
relies on a number of pragmatic observations
from which we derived
specific accelerations, for each sub-step of the optimization
(Secs. \ref{sec_fastPersistenceUpdate} and \ref{sec_fastDistance}).
\discuss{Our accelerations are simple and easy to implement, but result in
significant
gains in terms of runtimes.}
Extensive experiments
(\autoref{sec_timePerformance})
report
\discuss{$\times 60$}
accelerations
on average over state-of-the-art
frameworks (with both fewer and faster iterations),
thereby
making topological simplification optimization practical for real-life datasets.
We
illustrate
the utility of our contributions in two applications. First,
our work enables the direct visualization and analysis of topologically
simplified data (\autoref{sec_dataSimplification}). This reduces visual clutter
in isosurfaces by simplifying their topology (fewer components and handles). We
also
investigate
filament extraction in
three-dimensional
data, where
we show that our
approach helps standard topological techniques for
removing filament loops.
Second,
we show how
to use our approach to
repair genus defects in surface  processing
(\autoref{sec_genusReparation}).

%% file: relatedWork.tex
\subsection{Related work}
\label{sec_relatedWork}

Beyond post-process simplification schemes tailored for specific topological
abstractions (e.g. for merge/contour trees \cite{carr_phd04}, Reeb graphs
\cite{pascucci07} or Morse-Smale complexes \cite{gyulassy_phd08, GyulassyBHP11,
GuntherRSW14}), the
literature related to pre-process simplification can be classified into two
categories.

\noindent
\textbf{Combinatorial methods:} the first combinatorial approach for the
topological simplification of scalar data on surfaces has been proposed by
Edelsbrunner et al. \cite{EdelsbrunnerMP06}. This work can be seen as a
generalization of previous approaches in terrain modeling where only
persistence pairs involving minima were removed \cite{soille04, AgarwalAY06}.
Attali et al. \cite{attali09} extended this framework to generic filtrations,
while Bauer et al. \cite{BauerLW12} extended it to discrete Morse theory
\cite{forman98}. Tierny et al. presented a generalized approach
\cite{tierny_vis12}, supporting a variety of simplification criteria, which was
later extended by Lukasczyk et al. \cite{Lukasczyk_vis20} with an efficient
shared-memory parallel algorithm.
Such combinatorial simplification algorithms can be  used directly within
optimization
procedures \cite{Nigmetov24}, to remove noise in the solution at each
iteration.
While most of the above approaches were
specifically designed for scalar data on surfaces, they can be directly applied
to domains of higher dimensions. However, they can only simplify persistence
pairs involving extrema. For instance, this means that they cannot remove saddle
pairs in three-dimensional scalar fields, thus preventing
an advanced
simplification of this type of datasets.
\revision{\emph{Optimal simplification} \cite{BauerLW12} of scalar data is a
more general variant of the \emph{sublevel set simplification} problem, itself
being NP-hard in 3D \cite{AttaliBDGL13}.}
Then,
a polynomial time algorithm solving this problem may not
even exist. This theoretical limitation requires a shift in
strategy.

\noindent
\textbf{Numerical methods:} in contrast to combinatorial methods, which come
with strong guarantees on the result, numerical approaches
aim at providing
an approximate solution in a \discuss{best effort} manner. In other words,
these methods may not \emph{fully} simplify three-dimensional scalar fields up
to the desired tolerance either, but they will do their best to provide a
result as close as possible to the specified simplification. As such,
this type of approaches appear as a practical alternative overcoming the
theoretical
limitation
of combinatorial approaches discussed above.
In geometric modeling, several techniques have been described to generate
smooth scalar fields on surfaces, with a minimal number of critical points
\cite{NiGH04, GingoldZ07, PataneF09}.
Bremer et al. \cite{BremerEHP04} proposed a method based on Laplacian smoothing
to reconstruct a two-dimensional scalar field corresponding to a pre-simplified
Morse-Smale complex.
This work
has been extended by Weinkauf et al. \cite{WeinkaufGS10}
to bi-Laplacian optimization,
with an additional
enforcement of gradient continuity across the separatrices of the Morse-Smale
complex. While an extension of this work has been documented for the 3D case
\cite{GuntherJRSSW14}, it only addresses the simplification of persistence
pairs involving extrema, without explicit control on the saddle pairs.
Recently, a new class of methods dedicated to \emph{persistence optimization}
has been documented. Specifically, these approaches introduce a framework for
optimizing a dataset,
according to criteria expressed with persistence diagrams,
with
applications in various tasks including surface matching
\cite{PoulenardSO18}, point cloud processing \cite{GabrielssonGSG20},
classification \cite{CarriereCGIKU21} and more.
Solomon et al.
\cite{SolomonWB21} presented an approach based on stochastic subsampling
applied to 2D images.
Carriere et al. \cite{CarriereCGIKU21} presented an efficient and generic
persistence
optimization framework, supporting a wide range of criteria and applications,
exploiting the convergence properties of stochastic sub-gradient descent
\cite{KingmaB14} for tame functions  \cite{DavisDKL20}.
Nigmetov et al.
\cite{Nigmetov22} presented an alternative method, drastically reducing the
number of optimization iterations, but at the cost of
significantly more computationally
expensive steps.
As described in \autoref{sec_approach}, one can leverage these
frameworks for the problem of topological simplification, however, with
impractical runtimes  for three-dimensional datasets of standard size (e.g.
\discuss{up to days of computation}). We address this issue in this work by
proposing
a practical approach for topological simplification optimization, with
substantial
accelerations
over
state-of-the-art
frameworks for persistence optimization \cite{CarriereCGIKU21}.

%% file: contributions.tex
\subsection{Contributions}
\label{sec_contributions}

This paper makes the following new contributions:
\begin{enumerate}[itemsep=0pt]
 \item \textbf{Algorithm:} We introduce a practical solver for the optimization
of
topological simplification for scalar data (\autoref{sec_algorithm}). Our
algorithm is based on two
 accelerations, which are tailored to the specific problem of
topological simplification:
  \begin{itemize}[leftmargin=3.75mm]
    \item We present a simple and practical procedure for the fast update of the
persistence diagram of the data along the optimization
(\autoref{sec_fastPersistenceUpdate}),
hence preventing a full re-computation at each step.
    \item We describe a simple and practical procedure for the fast update of
the pair
assignments between the diagram specified as target, and the persistence
diagram of the
optimized data (\autoref{sec_fastDistance}), also preventing a full
re-computation at each step.
  \end{itemize}
Overall, the combination of these accelerations makes topological
simplification optimization
tractable for real-life datasets.
 \item \textbf{Applications:} Thanks to its practical time performance, our
work sets up ready-to-use foundations for several concrete applications:
    \begin{itemize}[leftmargin=3.75mm]
    \item \emph{\discuss{Visualization of
    topologically simplified data}
(\autoref{sec_dataSimplification}):}
we illustrate the utility of our framework for the direct visualization and
analysis
of topologically
simplified  data.
Our approach reduces visual clutter in isosurfaces by simplifying
connected components as well as, in contrast to previous work, surface handles.
We also investigate prominent filament extraction in
3D
data, where we show that our approach helps standard topological techniques for
removing filament loops.
      \item \emph{Surface genus repair (\autoref{sec_genusReparation}):} we show
how to use our framework
to
repair
genus defects in surface processing,
\discuss{with an explicit  control on the employed
primitives
(cutting or
filling).}
    \end{itemize}

 \item \textbf{Implementation:}
 We provide a C++ implementation of our algorithm that can be used for
reproducibility purposes.

\end{enumerate}

%% file: background.tex
\section{Preliminaries}
\label{sec_background}
This section presents the
background to our work. We refer the
reader to textbooks \cite{edelsbrunner09,zomorodianBook} for
introductions to computational topology.

\subsection{Input data}
\label{sec_inputData}
The input data is provided as a piecewise-linear (PL) scalar field $f : \domain
\rightarrow \mathbb{R}$ defined on a $d$-dimensional simplicial complex
$\domain$ (with $d \leq 3$ in our applications).
If the data is provided on a regular grid, we consider for $\domain$ the
implicit Freudenthal
triangulation of the grid \cite{freudenthal42, kuhn60}.
In practice, the data values are
defined on the $\numberOfVertices$ vertices of $\domain$, in the form of a
\emph{data vector}, noted $\dataVector_f \in \mathbb{R}^{\numberOfVertices}$.
$f$ is assumed to be injective on the vertices (i.e., the
entries of $\dataVector_f$ are all distinct), which can be easily obtained in
practice via a variant of simulation of simplicity \cite{edelsbrunner90}.

\begin{figure}
  \includegraphics[width=\linewidth]{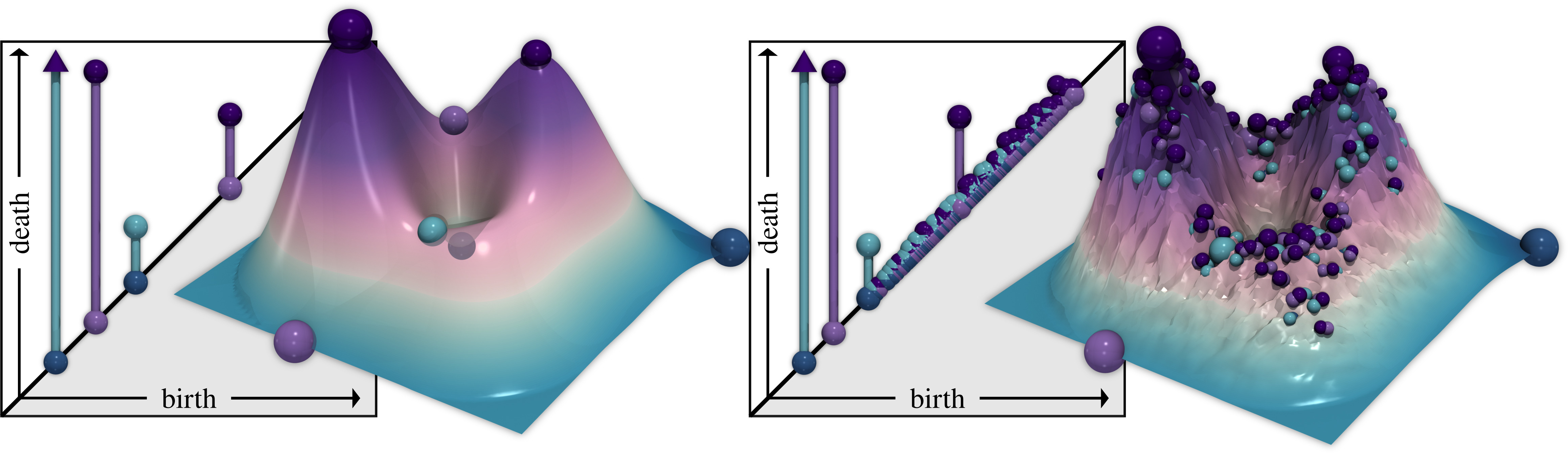}
  \caption{Persistent diagrams for the lexicographic filtration of a clean
(left) and a noisy (right)
terrain
example.
Minimum-saddle persistence pairs are show with cyan bars in the birth-death
space, while
saddle-maximum pairs are shown with purple bars.
Generators with
infinite persistence are marked with an upward arrow. The persistence of each
topological feature is given by the height of its bar.
Critical simplices are shown in the data with spheres, with a radius
proportional to
their
persistence.
}
  \label{fig_persistenceDiagram}
\end{figure}

\subsection{Persistence diagrams}
\label{sec_persistenceDiagrams}

Persistent homology has been developed independently by several research
groups \cite{B94, frosini99, robins99,
edelsbrunner02}.
Intuitively, persistent homology considers a sweep of the
data (i.e., a \emph{filtration}) and estimates at each step the
corresponding topological features (i.e., \emph{homology generators}), as well
as maps to the features of the previous step. This enables the identification of
the topological features, along with their lifespan, during the sweep.

In this work we consider the \emph{lexicographic filtration} (as described in
\cite{guillou_tvcg23}), which we briefly recall here for completeness.
Given the input data vector $\dataVector_f \in \mathbb{R}^{\numberOfVertices}$,
one can sort the vertices of $\domain$ by increasing data values, yielding a
\emph{global vertex order}. Based on this order, each $d'$-simplex $\simplex
\in \domain$ (with $d' \in [0, d]$) can be represented by the sorted list (in
decreasing values) of the $(d'+1)$ indices in the global vertex order of its
$(d'+1)$ vertices. Given this simplex representation, one can now compare two
simplices $\simplex_i$ and $\simplex_j$
via simple lexicographic
comparison, which induces a \emph{global lexicographic order} on the simplices
of $\domain$.
This order
induces a nested
sequence of simplicial complexes $\emptyset =  \domain_0 \subset \domain_1
\subset \dots \subset \domain_{\numberOfSimplices} = \domain$ (where
$\numberOfSimplices$ is the number of simplices of $\domain$), which we call
the \emph{lexicographic filtration} of $\domain$ by $f$.

At each step $i$ of the filtration, one can characterize the
$p^{th}$ homology group of $\domain_i$, noted
$\homologyGroup_p(\domain_i)$, for instance by
counting
its number of \emph{homology
classes} \cite{edelsbrunner09, guillou_tvcg23} (i.e., the \emph{order} of the
group) or its number of \emph{homology generators} (i.e., the \emph{rank} of
the group,
a.k.a. the
$p^{th}$
\emph{Betti number}, noted $\bettiNumber_p$).
Intuitively, in 3D, the first three Betti numbers ($\bettiNumber_0$,
$\bettiNumber_1$, and $\bettiNumber_2$) respectively provide the number of
connected components, of independent cycles and voids of the complex
$\domain_i$.
For two consecutive steps
of the filtration $i$ and $j$, the corresponding simplicial complexes are
nested ($\domain_i \subset \domain_j$). This inclusion induces
homomorphims between the homology groups $\homologyGroup_p(\domain_i)$ and
$\homologyGroup_p(\domain_j)$, mapping homology classes at step $i$ to homology
classes at step $j$. Intuitively, for the $0^{th}$ homology group, one can
precisely map a connected component at step $i$ to a connected component at step
$j$ because the former is included in the latter. In general,
a $p$-dimensional homology class $\gamma_i$ at step $i$ can be mapped to a
class $\gamma_j$ at step $j$ if the $p$-cycles of $\gamma_i$ and $\gamma_j$ are
\emph{homologous} in $\domain_j$ \cite{edelsbrunner09, guillou_tvcg23}. Then,
one can precisely track the homology generators between consecutive steps of the
filtration. In particular, a \emph{persistent generator} is born at step $j$
(with $j = i +1$) if it is not the image of any generator by the homomorphims
mapping $\homologyGroup_p(\domain_i)$  to $\homologyGroup_p(\domain_j)$.
Symmetrically, a persistent generator dies at step $j$ if it merges with
another, \emph{older} homology class, which was born before it (this is
sometimes
called the \emph{Elder rule} \cite{edelsbrunner09}). Each $p$-dimensional
persistent generator is associated to a
\emph{persistence pair}
$(\simplex_b, \simplex_d)$, where $\simplex_b$ is the $p$-simplex introduced at
the \emph{birth} of the generator (at step $b$) and where $\simplex_d$ is the
$(p+1)$-simplex introduced at its \emph{death} (at step $d$).
A $p$-simplex which is involved in the birth or the death of a generator is
called a \emph{critical simplex} and, in 3D, we call it a minimum, a
$1$-saddle,
a $2$-saddle, or a maximum if $p$ equals $0$, $1$, $2$, or $3$
\cite{robins_pami11, guillou_tvcg23}\revision{, respectively}.
The \emph{persistence} of the pair $(\simplex_b, \simplex_d)$, noted
$\persistence(\simplex_b, \simplex_d)$, is given by $\persistence(\simplex_b,
\simplex_d) = \dataVector_f(v_d) - \dataVector_f(v_b)$, where
$v_b$ and $v_d$ (the \emph{birth} and \emph{death vertices} of the pair) are the
vertices with highest global vertex order of $\simplex_b$ and $\simplex_d$.
We call \emph{zero-persistence pairs} the pairs
\revision{with}
$v_b
= v_d$.
\revision{Some} $p$-simplices of $\domain$ may be involved in no
persistence pair. These mark the birth of persistent generators with
\emph{infinite persistence} (i.e., which never die during the filtration) and
they characterize the homology groups of the final step of the filtration
($\domain_{\numberOfSimplices} = \domain$).

\begin{figure}
  
\includegraphics[width=\linewidth]{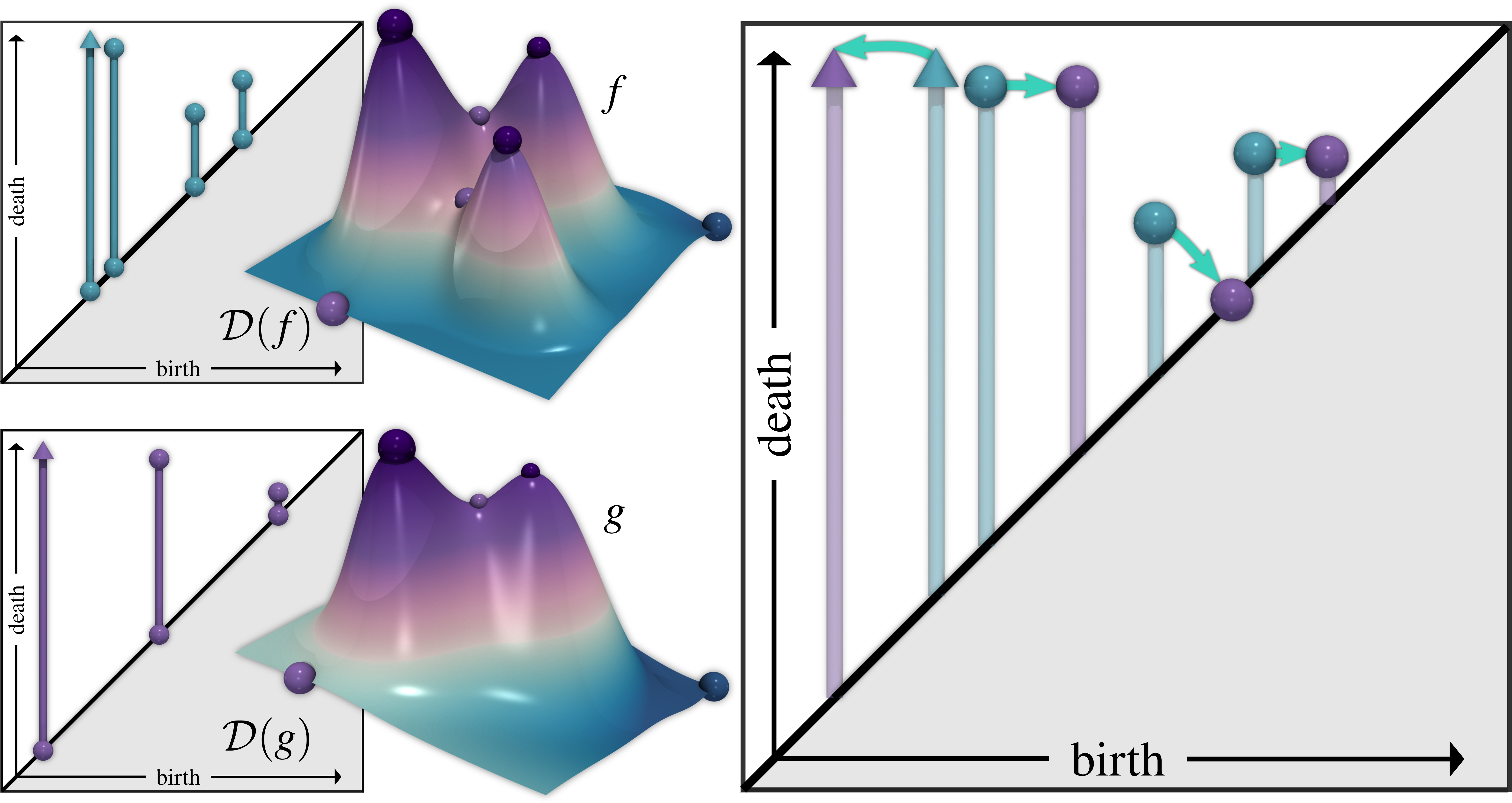}
  \caption{The Wasserstein distance $\wasserstein{2}$ between $\diagram(f)$
(top) and $\diagram(g)$ (bottom) is computed by assignment
optimization (\autoref{eq_wasserstein}) in the 2D birth-death space (right).
The optimal assignment $\phi^*$  (arrows) encodes a minimum cost transformation
of $\diagram(f)$ into $\diagram(g)$, which displaces persistence pairs in the
birth-death space or \emph{cancel} them by sending them to the diagonal.}
  \label{fig_wassersteinDistance}
\end{figure}

The set of persistence pairs induced by the lexicographic filtration of
$\domain$ by $f$ can be organized in a concise representation called the
\emph{persistence diagram} (\autoref{fig_persistenceDiagram}), noted
$\diagram(f)$, which embeds each non zero-persistence  pair $(\simplex_b,
\simplex_d)$ as a point in a 2D space (called the birth-death space), at
coordinates $\big(\dataVector_f(v_b), \dataVector_f(v_d)\big)$. By convention,
generators with infinite persistence are reported at coordinates
$\big(\dataVector_f(v_b), \dataVector_f(v_{max})\big)$, where $v_{max}$ is the
last vertex in the global vertex order.

\begin{figure*}
  \includegraphics[width=\linewidth]{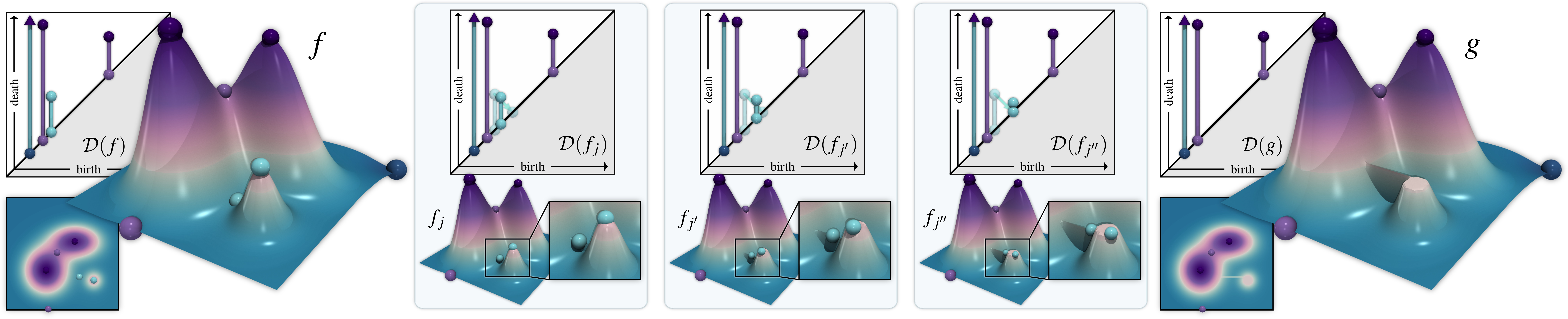}
  \caption{Optimizing the simplification of an input scalar field $f = f_0$ into
a
field $g = f_{final}$ for the removal of a user selected saddle-maximum pair
(cyan).
At each iteration ($j < j' < j''$), given the point $p_i \in \diagram(f_j)$
to cancel, the data values of its
birth and death
vertices
$v_{i_b}$ and
$v_{i_d}$ (cyan spheres in the data) are modified
to project $p_i$ to the diagonal.
In this example, this results in
a
scalar field $g$
which is close to $f$,
with the prescribed topology.
}
  \label{fig_optimization}
\end{figure*}

\subsection{Wasserstein distance between persistence diagrams}
\label{sec_wasserstein}

Two diagrams $\diagram(f)$ and $\diagram(g)$ can be reliably compared in
practice with the notion of \emph{Wasserstein distance}. For this, the two
diagrams $\diagram(f)$ and $\diagram(g)$ need to undergo an \emph{augmentation}
pre-processing phase. This step ensures that the two diagrams admit the same
number of points, which will facilitate their comparison.
Given a point $p = (p_b, p_d) \in \diagram(f)$, we note
$\projection(p)$ its diagonal projection: $\projection(p) =
\big({{1}\over{2}}(p_b + p_d), {{1}\over{2}}(p_b
+ p_d)\big)$. Let $\projection_f$ and $\projection_g$ be the sets of the
diagonal projections of the points of $\diagram(f)$ and
$\diagram(g)$ respectively. Then, $\diagram(f)$ and $\diagram(g)$ are
\emph{augmented} by appending to them the set of diagonal points $\projection_g$
and $\projection_f$ respectively. After this augmentation, we have
$|\diagram(f)| = |\diagram(g)|$.

Then, given two augmented persistence diagrams $\diagram(f)$ and $\diagram(g)$,
the $L^q$ Wasserstein distance between them is defined as:
\vspace{-1.25ex}
\begin{eqnarray}
\label{eq_wasserstein}
\wasserstein{q}\big(\diagram(f), \diagram(g)\big) = \min_{\phi \in \Phi}
\Big(\sum_{p \in \diagram(f)} c\big(p, \phi(p)\big)^q\Big)^{{{1}\over{q}}},
\end{eqnarray}

\vspace{-1.25ex}
\noindent
where $\Phi$ is the set of all bijective maps between the augmented diagrams
$\diagram(f)$ and $\diagram(g)$, which specifically map points of finite
(respectively infinite) $r$-dimensional persistent generators to points of
finite (respectively infinite) $r$-dimensional persistent generators. For this
distance, the cost $c(p, p')$ is set to zero when both $p$ and $p'$ lie on
the diagonal (i.e., matching dummy features has no impact on the distance).
Otherwise, it is set to the Euclidean distance in birth-death space
$||p - p'||_2$.

The Wasserstein distance
induces an
optimal assignment $\phi^*$ from  $\diagram(f)$ to  $\diagram(g)$
(\autoref{fig_wassersteinDistance}),  which depicts how to minimally transform
$\diagram(f)$ into $\diagram(g)$ (given the considered cost).
This
transformation may induce point displacements in the birth-death space, as well
as projections to the diagonal (encoding the \emph{cancellation} of a
persistence
pair).

\subsection{Persistence optimization}
\label{sec_persistenceOptimization}
Several frameworks have been
introduced for persistence optimization (\autoref{sec_relatedWork}). We review
a recent, efficient, and generic framework
\cite{CarriereCGIKU21}.

Given a scalar data vector $\dataVector_f \in \mathbb{R}^{\numberOfVertices}$
(\autoref{sec_inputData}), the purpose of persistence optimization is to
modify $\dataVector_f$ such that its persistence diagram
$\diagram(f)$ minimizes a certain loss $\loss$, specific to the
considered problem. Then the solution space of the optimization problem
is $\mathbb{R}^{\numberOfVertices}$.

Let $\filtration : \mathbb{R}^{\numberOfVertices} \rightarrow
\mathbb{R}^{\numberOfSimplices}$ be the \emph{filtration map}, which maps a
data vector $\dataVector_f$ from the solution space
$\mathbb{R}^{\numberOfVertices}$ to a filtration represented as a vector
$\filtration(\dataVector_f) \in \mathbb{R}^{\numberOfSimplices}$, where the
$i^{th}$ entry contains the index of the $i^{th}$ simplex $\simplex_i$ of
$\domain$ in the global lexicographic order (\autoref{sec_persistenceDiagrams}).
For convenience, we maintain a \emph{backward filtration map} $\filtration^{+} :
\mathbb{R}^{\numberOfSimplices} \rightarrow
\mathbb{R}^{\numberOfSimplices}$, which maps a filtration vector
$\filtration(\dataVector_f)$ to a vector in $\mathbb{R}^{\numberOfSimplices}$,
whose $i^{th}$ entry contains the index of the highest vertex (in global vertex
order) of the $i^{th}$ simplex in the global lexicographic order.

Given a persistence diagram $\diagram(f)$, the \emph{critical simplex
persistence order} can be introduced as follows. First,
the points of
$\diagram(f)$ are sorted by increasing birth and then, in case of birth ties, by
increasing death. Let us call this order the \emph{diagram order}.
Then the set of persistence pairs can also be sorted according to the diagram
order, by interleaving the birth and death simplices corresponding to each
point. This results in an ordering of the critical simplices, called the
\emph{critical simplex
persistence order}, where the $(2i)^{th}$ and $(2i+1)^{th}$ entries correspond
respectively to the birth and death simplices of the $i^{th}$ point $p_i$ in
the diagram order.
Critical simplices which are not
involved in a persistence pair (i.e., corresponding to homology classes of
infinite persistence) are appended to this ordering, in increasing order of
birth values.

Let us now consider the \emph{persistence map} $\persistenceMap:
\mathbb{R}^{\numberOfSimplices} \rightarrow \mathbb{R}^{\numberOfSimplices}$,
which maps a filtration vector $\filtration(\dataVector_f)$ to a persistence
image $\persistenceMap\big(\filtration(\dataVector_f)\big)$, whose $i^{th}$
entry contains the \emph{critical simplex persistence order} (defined above)
for the  $i^{th}$ simplex in the global lexicographic order. For
convenience, the entries corresponding to filtration indices which do not
involve
critical simplices are set to $-1$.

Now, to evaluate the relevance of a given diagram $\diagram(f)$ for the
considered optimization problem, one needs to define a loss term.
Let $\energy: \mathbb{R}^{\numberOfSimplices} \rightarrow
\mathbb{R}$ be an energy function, which evaluates the diagram energy given its
critical simplex persistence order. Then, given an input data vector
$\dataVector_f$, the associated loss $\loss : \mathbb{R}^{\numberOfVertices}
\rightarrow \mathbb{R}$ is given by:
\vspace{-1ex}
\begin{eqnarray}
\nonumber
\loss(\dataVector_f) = \energy \circ \persistenceMap \circ
\filtration(\dataVector_f).
\end{eqnarray}

\vspace{-1ex}
\revision{Since distinct functions can admit the same persistence
diagram, the global minimizer of the above loss
may not be unique. However, given the search space
($\mathbb{R}^{\numberOfVertices}$), the search for a global minimizer is 
still not
tractable in practice and local minimizers will be searched instead.}

If $\energy$ is locally Lipschitz
and a definable function of persistence, then
the composition $\energy \circ \persistenceMap \circ \filtration$ is also
definable and locally Lipschitz \cite{CarriereCGIKU21}. This implies that
\revision{the generic loss}
$\loss$ is differentiable almost everywhere and admits a well defined
sub-differential.
Then,
a
stochastic sub-gradient descent algorithm \cite{KingmaB14} converges almost
surely to a critical point of $\loss$ \cite{DavisDKL20}. In practice, this
means that the loss can be decreased by displacing
each diagram point $p_i$ in
the diagram $\diagram(f)$ according to the sub-gradient.
Assuming a constant
global
lexicographic order, this displacement
can be back-propagated into modifications in data values in the
vector $\dataVector_f$, by identifying the vertices $v_{i_b}$ and $v_{i_d}$
corresponding to the birth and death (\autoref{sec_persistenceDiagrams}) of the
$i^{th}$ point in the diagram order:
\vspace{-1ex}
\begin{equation}
\begin{array}{r@{}l}
\label{eq_diagramToVertex}
v_{i_b} &{}= \filtration^{+}\big(\persistenceMap^{-1}(2i)\big)\\
v_{i_d} &{}= \filtration^{+}\big(\persistenceMap^{-1}(2i + 1)\big),
\end{array}
\end{equation}

\vspace{-1ex}
\noindent
and by updating their data values $\dataVector_f(v_{i_b})$ and
$\dataVector_f(v_{i_d})$ accordingly.

%% file: approach.tex
\section{Approach}
\label{sec_approach}
This section describes our \discuss{overall} approach for the
optimization of
the topological simplification of scalar data.

Given the diagram $\diagram(f)$ of the input field $f$, we call a \emph{signal}
pair a persistence pair of $\diagram(f)$ which is selected by the user for
preservation. Symmetrically, we call a \emph{non-signal} pair a persistence pair
of $\diagram(f)$ which is selected by the user for cancellation.
Note that this distinction between \emph{signal} and \emph{non-signal} pairs is
application dependent.
In practice, the user can be aided by
several criteria, such as
persistence \cite{edelsbrunner02},
\revision{geometric}
measures \cite{carr04}, etc.
Then, topological simplification can be expressed as an optimization problem,
with the following objectives:
\begin{enumerate}
\item Penalizing the persistence of the \emph{non-signal} pairs;
\item Enforcing the precise preservation of the \emph{signal} pairs.
\end{enumerate}

In short, we wish to penalize the undesired  features
(objective \emph{1}), and, at the same time, enforce the precise preservation
of the
features of the input which are deemed relevant (objective \emph{2}). The latter
objective is important in practice to preserve the  accuracy of the
features of interest. As later discussed in \autoref{sec_fastDistance} (and
illustrated in \autoref{fig_stillPairs}), \emph{non-signal} and \emph{signal}
pairs do interact during the optimization, thereby perturbing \emph{signal}
pairs.
In our experiments (\autoref{sec_results}),
at each optimization iteration,
\discuss{$11\%$} of the \emph{signal} pairs are perturbed by \emph{non-signal}
pairs (on average, and up to \discuss{$32\%$}).
In certain configurations, this can drastically alter the persistence
of \emph{signal} pairs. Hence, to address this issue, the precise preservation
of the \emph{signal} pairs should be explicitly
constrained.

In the following, we formalize this specific optimization problem based on the
generic framework described in \autoref{sec_persistenceOptimization}.
Our novel solver (for efficiently solving it) is
presented in \autoref{sec_algorithm}.

Let $\target$ be the \emph{target diagram}. It can be obtained by copying the
diagram $\diagram(f)$ of
the input field $f$, and by removing the \emph{non-signal} pairs.
$\target$ encodes the two objectives of our
problem: it describes
the constraints for the cancellation of the
noisy features of $f$ (objective \emph{1}) as well as
the lock constraints for its features of interest (objective \emph{2}).

In general,
a perfect
\emph{reconstruction} (i.e., a scalar field $g$\revision{, close to $f$,} such
that
$\diagram(g) = \target$) may not exist in 3D (deciding on its existence is
NP-hard
\cite{AttaliBDGL13}). Thus, a practical strategy consists in
optimizing
the scalar field $f$
such that its diagram $\diagram(f)$ gets
\emph{as close as possible} to $\target$ \revision{(and relaxing
$||f-g||_\infty$)}.
For this, we consider the following \emph{simplification energy} $\energy$
\revision{(to be used within the generic loss $\loss$, introduced in
\autoref{sec_persistenceOptimization})}:
\vspace{-1ex}
\begin{eqnarray}
  \label{eq_energy}
  \energy\big(\diagram(f)\big) = \wasserstein{2}\big(\diagram(f),
\target\big)^2.
\end{eqnarray}

\vspace{-1ex}
\noindent
Since the Wasserstein distance is locally
Lipschitz and a definable function of persistence \cite{CarriereCGIKU21}, the
optimization framework of \autoref{sec_persistenceOptimization}
can be used to optimize $\loss = \energy \circ
\persistenceMap \circ \filtration$ with guaranteed convergence.

Specifically, at each iteration, given the optimal assignment $\phi^*$ induced
by the Wasserstein distance between $\diagram(f)$ and $\target$
(\autoref{sec_wasserstein}), one can displace each point $p_i$ in $\diagram(f)$
towards its individual target $\phi^*(p_i)$ by adjusting accordingly the
corresponding scalar values $\dataVector_f(v_{i_b})$ and
$\dataVector_f(v_{i_d})$ (\autoref{eq_diagramToVertex}). In practice,
the generic
optimization framework reviewed in
\autoref{sec_persistenceOptimization} computes this displacement (given
$\phi^*$) via automatic differentiation \cite{CarriereCGIKU21} and
by using Adam \cite{KingmaB14} for gradient descent.

However, depending on the employed step size, a step of gradient descent on
$\dataVector_f$ may change the initial filtration order
(\autoref{sec_persistenceDiagrams}). Thus, after a step of gradient descent,
the persistence diagram of the optimized data needs to be recomputed and,
thus, so does its optimal assignment $\phi^*$ to the target $\target$.
This procedure is then iterated, until the loss at the current iteration is
lower than a user-specified fraction $s$ of the loss at the first iteration
(or until a maximum number $j_{max}$ of iterations).
We call this overall procedure the \emph{baseline optimization for topological
simplification}. It is summarized in
\autoref{alg_baseline} and illustrated
in
\autoref{fig_optimization}.

\begin{algorithm}[t]
    \small
    \caption{\small Baseline optimization approach
for topological simplification.}
\label{alg_baseline}
\hspace*{\algorithmicindent} \textbf{Input}: Input scalar field $f = f_0 :
\domain
\rightarrow \mathbb{R}$.

\hspace*{\algorithmicindent} \textbf{Input}: Target diagram $\target$.

\hspace*{\algorithmicindent} \textbf{Input}: Stopping conditions $s
\in [0, 1]$, $j_{max} \in \mathbb{N}$.

\hspace*{\algorithmicindent} \textbf{Output}: Topologically simplified scalar
field $g = f_{final} : \domain \rightarrow \mathbb{R}$.

\begin{algorithmic}[1]

\State $j \leftarrow 0$
  \State $\diagram(f_j)
\leftarrow PersistenceDiagramComputation(\dataVector_{f_j})$
  \State $\big(\loss(\dataVector_{f_j}), \phi^*_j\big) \leftarrow
WassersteinDistanceComputation\big(\diagram(f_j), \target\big)$

\Do
  \State $j \leftarrow j + 1$
  \State $\dataVector_{f_j} \leftarrow
GradientDescentStep\big(\phi^*_{j-1}, \dataVector_{f_{j-1}}\big)$

    \State $\diagram(f_j)
\leftarrow PersistenceDiagramComputation(\dataVector_{f_j})$
  \State $\big(\loss(\dataVector_{f_j}), \phi^*_j\big) \leftarrow
WassersteinDistanceComputation\big(\diagram(f_j), \target\big)$
\doWhile{$\loss\big(\dataVector_{f_j}\big) > s
\loss\big(\dataVector_{f_0}\big)$ and $j < j_{max}$}
\end{algorithmic}
\end{algorithm}

As shown in \autoref{alg_baseline}, each iteration $j$ of the optimization
involves a step of gradient descent, the computation of the diagram
$\diagram(f_j)$ and the computation of its Wasserstein distance to $\target$.
While the first of these three steps has linear time complexity, the other two
steps are notoriously computationally expensive and both have cubic
theoretical worst case time complexity, $\complexity(\numberOfSimplices^3)$. In
practice, practical implementations for persistence diagram computation tend to
exhibit a quadratic behavior \cite{BauerKRW17, dipha, guillou_tvcg23}.
Moreover, the exact optimal assignment algorithm \cite{Munkres1957} can be
approximated in practice to improve runtimes, for instance with Auction-based
 \cite{Bertsekas91,
Kerber2016, vidal_vis19} or sliced approximations \cite{CarriereCO17}.

However, even when using the above practical implementations for persistence 
computation and assignment optimization, 
the baseline optimization
approach for
topological simplification has impractical runtimes for datasets of standard
size.
Specifically, for the simplifications considered in our experiments
(\autoref{sec_results}),
this approach
can require up to days of computations per dataset.
When it completes within 24 hours, the computation spends
\discuss{$20 \%$} of the time in persistence computation and \discuss{$75 \%$}
in assignment optimization.

%% file: algorithm.tex
\section{Algorithm}
\label{sec_algorithm}

This section describes our algorithm for topological simplification
optimization. It is based on a number of practical accelerations of the
baseline optimization (\autoref{alg_baseline}), which are particularly relevant
for the problem of topological simplification.

\subsection{Direct gradient descent}
\label{sec_gradientDescent}
Instead of relying on automatic differentiation
and
the Adam optimizer \cite{KingmaB14}
as done in the generic
framework reviewed in \autoref{sec_persistenceOptimization}, similar to
\cite{PoulenardSO18}, we can derive the analytic expression of the gradient of
our energy \revision{on a per-iteration basis} (\autoref{eq_energy}) and
perform
\revision{at each iteration} a \revision{direct step of} gradient descent, in
order to improve performance.
\revision{Specifically, at}
the iteration $j$ (\autoref{alg_baseline}), given the
current persistence diagram $\diagram(f_j)$,
if the assignments between diagonal points are ignored (these have zero cost,
\autoref{sec_wasserstein}),
\autoref{eq_energy} can be
re-written as:
\vspace{-1ex}
\begin{eqnarray}
\nonumber
\energy\big(\diagram(f_j)\big) = \min_{\phi \in \Phi} \sum_{p_i \in
\diagram(f_j)} ||p_i - \phi(p_i)||_2^2.
\end{eqnarray}

\vspace{-1ex}
\revision{As the optimal assignment  $\phi^*_j$ (i.e.,
minimizing the
energy
for a
fixed $\diagram(f_j)$) is constant at the iteration $j$,}
the energy can be re-written as:
\vspace{-1ex}
\begin{eqnarray}
\nonumber
\energy\big(\diagram(f_j)\big)
= \sum_{p_i \in \diagram(f_j)}
\big(p_{i_b} - \phi^*_j(p_i)_b\big)^2 + \big(p_{i_d} -
\phi^*_j(p_i)_d\big)^2.
\end{eqnarray}

\vspace{-1ex}
Then,
given \autoref{eq_diagramToVertex},
\revision{for the iteration $j$,}
the overall optimization loss $\loss(\dataVector_{f_j})$
can be expressed as a function of the input data vector $\dataVector_{f_j}$:
\vspace{-1ex}
\begin{eqnarray}
\nonumber
\loss(\dataVector_{f_j})
= \sum_{p_i \in \diagram(f_j)}
\big(\dataVector_{f_j}(v_{i_b}) - \phi^*_j(p_i)_b\big)^2 +
\big(\dataVector_{f_j}(v_{i_d}) -
\phi^*_j(p_i)_d\big)^2.
\end{eqnarray}

\vspace{-1ex}
\noindent
\revision{Then, for the iteration $j$, given the constant assignment
$\phi^*_j$,}
this
\revision{loss}
is
convex with
$\dataVector_{f_j}$ \revision{(in addition to being
locally Lipschitz)} and gradient descent can be
considered. Specifically, let $\nabla v_{i_b} \in
\mathbb{R}^{\numberOfVertices}$ be a vector with
zero entries, except for the ${{i_b}}^{th}$ entry,
set to $1$.
Let $\nabla v_{i_d} \in \mathbb{R}^{\numberOfVertices}$ be the vector
constructed
similarly for ${v_{i_d}}$.
Then, by the chain rule,
we have:
\vspace{-1ex}
\begin{eqnarray}
 \nonumber
\nabla \loss(\dataVector_{f_j})
& = & \sum_{p_i \in \diagram(f_j)}
\Big(
2\big(\dataVector_{f_j}(v_{i_b}) - \phi^*_j(p_i)_b\big)  \nabla
v_{i_b}\\
\nonumber
& + &
2\big(\dataVector_{f_j}(v_{i_d}) -
\phi^*_j(p_i)_d\big) \nabla v_{i_d}\Big).
\end{eqnarray}

\vspace{-1ex}
\noindent
We now observe that the gradient can be split into two terms, a \emph{birth
gradient} (noted $\nabla \loss(\dataVector_{f_j})_b$) and a \emph{death
gradient} (noted $\nabla \loss(\dataVector_{f_j})_d$):
\vspace{-1ex}

\begin{equation}
\label{eq_birthAndDeathGradients}
\begin{array}{r@{}l}
\nonumber
\nabla \loss(\dataVector_{f_j})_b
&{}= \sum_{p_i \in \diagram(f_j)}
2\big(\dataVector_{f_j}(v_{i_b}) - \phi^*_j(p_i)_b\big)  \nabla
v_{i_b}\\
\nonumber
\nabla \loss(\dataVector_{f_j})_d
&{}=
\sum_{p_i \in \diagram(f_j)}
2\big(\dataVector_{f_j}(v_{i_d}) -
\phi^*_j(p_i)_d\big) \nabla v_{i_d}.
\end{array}
\end{equation}

\vspace{-1ex}
Then, given the above gradient expressions, a step of gradient descent is
obtained by:
\vspace{-1ex}
\begin{eqnarray}
\nonumber
 \dataVector_{f_{j+1}} = \dataVector_{f_j} - \big((\alpha_b  \nabla
\loss(\dataVector_{f_j})_b
+ \alpha_d  \nabla
\loss(\dataVector_{f_j})_d
\big),
\end{eqnarray}

\vspace{-1ex}
\noindent
where $\alpha_b, \alpha_d \in \mathbb{R}$ are
the gradient step sizes for the birth and death gradients respectively.
Such individual
step sizes enable an explicit control over the evolution of
the persistence pairs to cancel (see \autoref{sec_genusReparation}).

\subsection{Fast persistence update}
\label{sec_fastPersistenceUpdate}

\begin{figure}
\vspace{-1ex}
  \includegraphics[width=\linewidth]{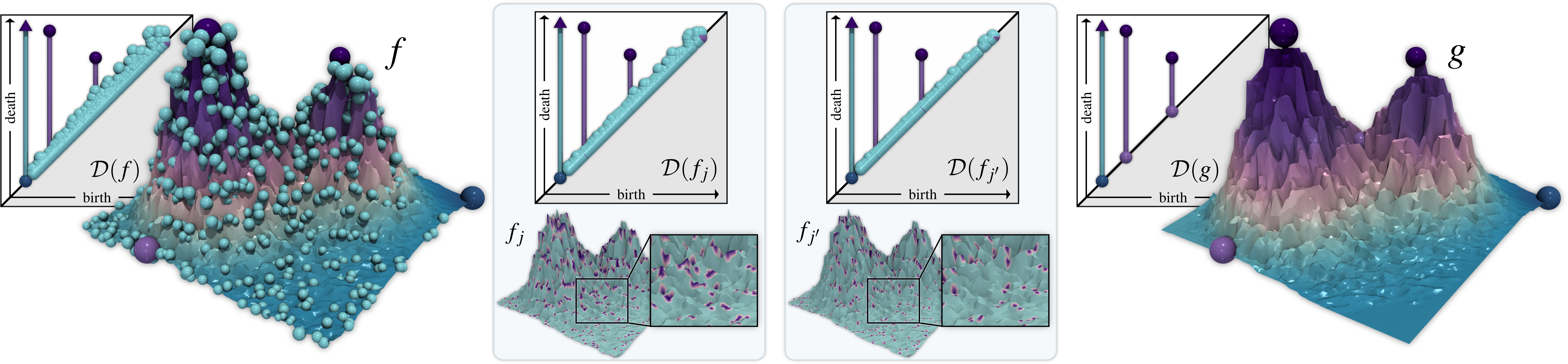}
 \caption{Updated vertices (dark purple vertices, center insets)
along the topological simplification
optimization of a noisy terrain (\emph{non-signal} pairs, to simplify, are shown
in cyan). In this
example, only 20\% of the
vertices are updated per iteration on average. The discrete gradient
(at the core of a recent, fast persistence computation algorithm
\cite{guillou_tvcg23}) only needs to be recomputed for these, yielding
a \discuss{x2} speedup for persistence computation.}
\label{fig_updatedVertices}
\end{figure}

As described in \autoref{sec_approach}, each optimization iteration $j$
involves the computation of the persistence diagram of the
data vector $\dataVector_{f_j}$, which is computationally expensive
(\discuss{$20 \%$} of the computation time on average).
Subsequently, for each persistence pair
$p_i$, the data values of its vertices $v_{i_b}$ and $v_{i_d}$ will be updated
 given the optimal assignment $\phi^*_j$.

A key observation
can be leveraged to improve the performance of the persistence computation
stage.
Specifically, the updated data vector $\dataVector_{f_{j+1}}$ only contains
updated
data values for the subset of the vertices of $\domain$ which are the birth and
death vertices $v_{i_b}$ and $v_{i_d}$ of a persistence pair $p_i$.
Then,
only
a small fraction of the
vertices are updated from one iteration to the next,
as shown in \autoref{fig_updatedVertices}.
In practice,
for the simplifications considered in our experiments (\autoref{sec_results}),
\discuss{$90 \%$ of the vertices of $\domain$} do \emph{not} change
their data values between consecutive iterations (on average over our
datasets, with the baseline optimization).
This indicates that a procedure capable of quickly updating the persistence
diagram $\diagram(f_{j+1})$ based on $\diagram(f_{j})$ has the potential to
improve performance in practice.

Several approaches focus on updating a
persistence diagram based on a previous estimation \cite{CohenSteinerEM06,
Luo24},
with a time complexity that is linear
for each vertex order
transposition between the two scalar fields.
However, this number of
transpositions can be extremely large in practice.

 Instead, we derive a simple procedure
based on recent work for computing persistent homology with Discrete Morse
Theory (DMT) \cite{forman98, robins_pami11, guillou_tvcg23}, which we briefly
review here for completeness. Specifically,
the
\emph{Discrete Morse Sandwich}
(DMS) approach \cite{guillou_tvcg23}
revisits the seminal algorithm \emph{PairSimplices}
\cite{edelsbrunner02} within the DMT setting, with specific accelerations for
volume datasets. This algorithm is based on two main steps.
First, a \emph{discrete gradient field} is computed, for the fast
identification of zero-persistence pairs. Second, the remaining persistence
pairs are computed by restricting the algorithm
\emph{PairSimplices} to the critical simplices (with specific accelerations for
the
persistent
homology groups of dimension $0$ and $d-1$).

The first key practical insight about this algorithm is that its first step,
discrete gradient computation, is documented to represent in practice, in 3D,
$66\%$ of
the persistence computation time on average \cite{guillou_tvcg23} (in sequential
mode). This indicates that, if one could quickly update the discrete gradient
between consecutive iterations,
the overall persistence computation step could be accelerated
by up to a factor of $3$ in practice.

The second key practical insight about this algorithm is that the discrete
gradient computation is a completely \emph{local} operation, specifically to
the lower star of each vertex $v$ \cite{robins_pami11} (i.e., the co-faces of
$v$ containing no higher vertex than $v$ in the global vertex order).

Thus, we leverage the above two observations to expedite the computation of the
diagram $\diagram(f_{j})$, based on the diagram $\diagram(f_{j-1})$.
Specifically, we mark as \emph{updated} all the vertices
of $\domain$
for which the data
value is updated by gradient descent at
iteration $j-1$  (\autoref{sec_gradientDescent}). Then, the discrete gradient
field at step $j$ is copied from that at step $j-1$ and the local discrete
gradient computation procedure \cite{robins_pami11} is only re-executed
for these vertices for which the lower star may have changed from step $j-1$ to
step $j$,
i.e. the vertices marked as \emph{updated} or which contain
\emph{updated} vertices in their star.
This localized update guarantees
the computation of the correct discrete gradient field at step $j$, with a
very small
number of local re-computations. Next, the second step of the
DMS algorithm
\cite{guillou_tvcg23} (i.e., the
computation of the persistence pairs from the critical simplices) is re-executed
as-is.

\begin{figure}
  \includegraphics[width=\linewidth]{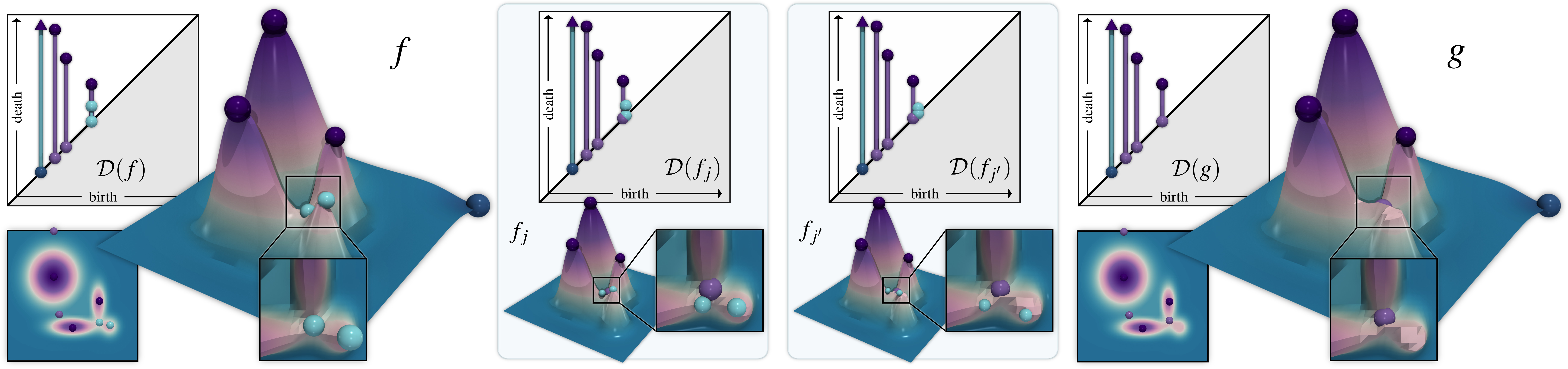}
  \caption{Interactions between \emph{non-signal} and \emph{signal} pairs
during the optimization. A multi-saddle vertex can be involved in
both a \emph{non-signal} pair $p$ (cyan bar in $\diagram(f)$) and
a \emph{signal} pair $p'$ (vertically aligned purple bar in $\diagram(f)$). At
iteration $j$, the update of the \emph{non-signal} pair $p$ unfolds the
multi-saddle into multiple simple saddles of distinct values, effectively
\emph{perturbing} the birth of the \emph{signal} pair $p'$ and making it
\emph{non-still}.
In real-life data, especially in 3D, such configurations  occur  often, and
cascade.
In our experiments (\autoref{sec_results}), at each iteration, \discuss{$11\%$}
of the \emph{signal} pairs are perturbed this way by \emph{non-signal} pairs
(on average, and up to \discuss{$32\%$}).
This is addressed by our loss (\autoref{sec_approach}) which
enforces \emph{signal} pair preservation.}

\label{fig_stillPairs}
\end{figure}

\subsection{Fast assignment update}
\label{sec_fastDistance}

As described in \autoref{sec_approach},
each optimization iteration $j$
involves the computation of the optimal assignment $\phi^*_j$ from the current
diagram $\diagram(f_j)$ to the target $\target$, which is computationally
expensive.

However, for the problem of simplification, a key
practical observation can be leveraged to accelerate this assignment
computation.
In practice, an important fraction of the pairs of $\diagram(f_j)$ to
optimize (among \emph{signal} and \emph{non-signal} pairs) may only move
slightly in the domain from one iteration to the next (as illustrated in
\autoref{fig_stillPairs}), and some do not move at all.
For these pairs which do not move at step $j$, the assignment
can be re-used from the step $j-1$, hence reducing the size of the
 assignment problem
(\autoref{sec_wasserstein}), and hence
reducing its practical runtime.

Given two persistence diagrams $\diagram(f_j)$ and $\diagram(f_{j-1})$,
we call a \emph{still} persistence pair a pair of points $(p_i, p'_i)$ with $p_i
\in \diagram(f_j)$ and $p'_i \in \diagram(f_{j-1})$ such that $v_{i_b} =
v_{i'_b}$ and $v_{i_d} = v_{i'_d}$. In other words, a \emph{still} persistence
pair is a pair which does not change its birth and death vertices from one
optimization iteration to the next. In practice,
for the simplifications considered in our experiments (\autoref{sec_results}),
\discuss{$84\%$} of the persistence pairs of
$\diagram(f_j)$ are still (on average over the iterations and our test
datasets, \autoref{sec_results}). This indicates that a substantial speedup
could be obtained by expediting the assignment computation for still pairs.

Let $\still$ be the set of still pairs between the iteration $j$ and $j-1$.
Then, for each pair $(p_i, p'_i) \in \still$, we set $\phi^*_j(p_i) \leftarrow
\phi^*_{j-1}(p_i')$. Concretely, we re-use at step $j$
the assignment at step $j-1$
for all the still
pairs.

Next, let $\overline{\diagram(f_j)}$ be the \emph{reduced}
diagram at step $j$, i.e., the subset of $\diagram(f_j)$ which does not contain
still pairs: $\overline{\diagram(f_j)} = \diagram(f_j) - \{p_i \in
\diagram(f_j), (p_i, p'_i) \in
\still\}$. Similarly,
let $\overline{{\target}_j}$ be the \emph{reduced} target at step $j$, i.e.,
the subset of $\target$ which has not been assigned to still pairs:
$\overline{{\target}_j} = \target - \{p_i'' \in \target,
p_i'' = \phi^*_{j}(p_i),
(p_i, p'_i) \in
\still\}$.
Then, we
finally
complete
the assignment between
$\diagram(f_j)$ and $\target$ by computing the Wasserstein distance between
$\overline{\diagram(f_j)}$ and $\overline{{\target}_j}$, as documented in
\autoref{sec_wasserstein}.

Note that, in the special case where the reduced target $\overline{{\target}_j}$
is empty (i.e.,
all \emph{signal} pairs
are \emph{still}), the reduced diagram $\overline{\diagram(f_j)}$ only contains
\emph{non-signal} pairs. Then,
the
optimal assignment can be readily obtained
(without any assignment
optimization)
by simply assigning each point $p_i$ in
$\overline{\diagram(f_j)}$
to
its diagonal projection $\projection(p_i)$. However, from our experience, such
a perfect scenario never occurs on real-life data, at the notable exception of
the very first iteration (before the data values are
actually
modified by the
solver). For the following iterations, many \emph{signal} pairs are not still
in practice.
\autoref{fig_stillPairs} illustrates this with a simple 2D example involving a
multi-saddle vertex.
However, in real-life data, such configurations
occur very often, and
cascade.
Also,
these configurations get significantly more challenging
in 3D. For instance,
the birth and death vertices of a given \emph{signal} pair can both be
multi-saddles, themselves possibly involved with \emph{non-signal} pairs to
update (hence yielding perturbations in the \emph{signal} pair).
In certain configurations, this can drastically alter the
persistence of the \emph{signal} pairs affected by such perturbations.
This is
addressed by our loss (\autoref{sec_approach}) which
enforces the
preservation of the \emph{signal} pairs via assignment optimization.

%% file: results.tex
\section{Results}
\label{sec_results}

This section presents experimental results obtained on a computer with 
two Xeon CPUs (3.0 GHz, 2x8 cores, \discuss{64GB} of RAM).
We implemented our algorithm (\autoref{sec_algorithm}) in C++ (with OpenMP) as 
a module for TTK \cite{ttk17, ttk19}. We implemented the baseline optimization 
approach (\autoref{sec_approach}) by porting the original implementation by
Carriere et al. \cite{CarriereCGIKU21} from \discuss{TensorFlow/Gudhi}
\cite{tensorflow2015-whitepaper, gudhi} to PyTorch/TTK \cite{pytorch} and by 
applying it to the loss described in \autoref{sec_approach}.
\revision{We chose this approach as a baseline, since
its implementation is simple and publicly
available, and since it provides performances comparable to alternatives
\cite{Nigmetov22}.}
In our implementations,
we use the
DMS algorithm \cite{guillou_tvcg23} for persistence computation
(as it is reported to provide the best practical performance for scalar
data)
and the
Auction algorithm  \cite{Bertsekas91,
Kerber2016} for the core assignment optimization, with a relative precision of
$0.01$, as recommended in the literature \cite{Kerber2016}.
\revision{Persistence computation with DMS
\cite{guillou_tvcg23}
is the
only step of our approach which leverages parallelism (see \cite{guillou_tvcg23}
for a detailed performance
analysis).}
Experiments were performed on a selection of $10$ (simulated and acquired) 2D
and 3D datasets 
extracted from 
public repositories \cite{ttkData, openSciVisDataSets}, with an emphasis on 3D 
datasets containing large filament structures (and thus possibly, many
persistent saddle pairs). The 3D datasets were resampled to a common resolution 
($256^3$), to better observe runtime variations based on the input topological 
complexity. Moreover, for each dataset, the data values were normalized to the
interval $[0, 1]$, to facilitate parameter tuning across distinct datasets.

Our algorithm is subject to two meta-parameters:
\discuss{the gradient step sizes $\alpha_b$ and $\alpha_d$.}
To adjust them,
we selected as default values the ones which minimized the runtime
for
our test dataset with the largest diagram.
This
resulted in
$\alpha_b = \alpha_d = 0.5$
(which coincides,
given a persistence pair to cancel,
to a displacement 
of its birth and death vertices
halfway 
towards the other, in terms of function range). For the baseline optimization 
approach (\autoref{sec_approach}), we set the initial learning rate of Adam 
\cite{KingmaB14} to the largest value
which still
enabled practical convergence
for
all
our datasets (specifically, $10^{-4}$). For both approaches, we set the maximum
number of iterations $j_{max}$ to $1,000$ (however, it has never been reached
in our performance experiments).

\begin{table}
\vspace{-1.5ex}
\caption{Time performance comparison between the
baseline optimization approach (\autoref{sec_approach}) and our solver
(\autoref{sec_algorithm}), for
a basic simplification
(\emph{non-signal} pairs: input pairs less
persistent than $1\%$ of the function range).
\revision{The column N.S.S.P. reports the average percentage of non-still
\emph{signal} pairs
for our solver.}
The stopping condition
is set to $s = 0.01$.\vspace{-1ex}}
    \fontsize{7}{7}\selectfont
    \def\arraystretch{0.5}
    \setlength{\tabcolsep}{0.57em}
\adjustbox{width=\linewidth,center}{
    \begin{tabular}{|l|r|r|r||r|r|r||r|r|r|r|r|}
    \hline
    \multirow{2}{*}{Dataset}
    & \multirow{2}{*}{$d$}
    & \multirow{2}{*}{$|\diagram(f)|$}
    & \multirow{2}{*}{$|\target|$} &
    \multicolumn{3}{c||}{Baseline (\autoref{sec_approach})} &
    \multicolumn{5}{c|}{Our solver (\autoref{sec_algorithm})} \\
    ~
    & ~
    & ~
    & ~ &
        \#It.
        &
        \revision{Time/It (s.)}
        &
        Time (s.) &
        \revision{N.S.S.P.} & 
        \#It.
          &
          \revision{Time/It (s.)}
          &
        Time (s.)
        & Speedup
        \\
    \hhline{|=|=|=|=||=|=|=||=|=|=|=|=|}
      Cells  & 2 &
      $7,676$ &
      $2,635$ &
      $89$ &
      \revision{$0.58$} &
      $52$ &
      \revision{$0.07\%$} &
      $10$&
      \revision{$0.20$} &
      $2$& \textbf{26}\\
      \arrayrulecolor{lightgray}
      \hline
      Ocean Vortices & $2$ &
      $12,069$ &
      $2,781$ &
      $87$ &
      \revision{$0.61$} &
      $53$ &
      \revision{$8.02\%$} &
      $12$ &
      \revision{$0.25$} &
      $3$ &  \textbf{18}\\
     \arrayrulecolor{black}
      \hline
      Aneurysm & $3$ &
      $38,490$ &
      $24,725$ &
      $80$ &
      \revision{$29.89$} &
      $2,391$ &
      \revision{$3.70\%$} &
      $8$ &
      \revision{$11.63$} &
      $93$ &  \textbf{26}\\
      \arrayrulecolor{lightgray}
      \hline
      Bonsai & $3$ &
      $168,489$ &
      $55,464$ &
      $67$ &
      \revision{$56.73$} &
      $3,801$ &
      \revision{$3.90\%$} &
      $10$ &
      \revision{$13.50$} &
      $135$ &  \textbf{28}\\
      \arrayrulecolor{lightgray}
      \hline
      Foot & $3$ &
      $754,965$ &
      $474,271$ &
      $60$ & 
      \revision{$914.47$} &
      $54,868$ &
      \revision{$11.28\%$} &
      $4$ &
      \revision{$104.25$} &
      $417$ &  \textbf{{132}}\\
      \arrayrulecolor{lightgray}
      \hline
      Neocortical Layer Axon & $3$ &
      $765,406$ &
      $483,791$ &
      $89$ & 
      \revision{$735.36$} &
      $65,447$ &
      \revision{$32.04\%$} &
      $8$ &
      \revision{$263.38$} &
      $2,107$ &  \textbf{31}\\
      \arrayrulecolor{lightgray}
      \hline
      Dark Sky & $3$ &
      $1,140,653$ &
      $774,793$ &
      NA & 
      \revision{NA} &
      \textbf{> 24h} &
      \revision{$9.08\%$} &
      $6$ &
      \revision{$122.00$} &
      $732$ &  \textbf{> 118}\\
      \arrayrulecolor{lightgray}
      \hline
      Backpack & $3$ &
      $1,331,362$ &
      $84,402$ &
      $66$ & 
      \revision{$305.58$} &
      $20,168$ &
      \revision{$21.46\%$} &
      $9$ &
      \revision{$62.22$} &
      $560$ &  \textbf{36}\\
      \arrayrulecolor{lightgray}
      \hline
      Head Aneurysm & $3$ &
      $1,345,168$ &
      $234,672$ &
      NA & 
      \revision{NA} &
      \textbf{> 24h} &
      \revision{$6.68\%$} &
      $5$ &
      \revision{$83.80$} &
      $419$ &  \textbf{> 206}\\
      \arrayrulecolor{lightgray}
      \hline
      Chameleon & $3$ &
      $3,641,961$ &
      $32,578$ &
      $55$ & 
      \revision{$210.51$} &
      $11,578$ &
      \revision{$18.22\%$} &
      $8$ &
      \revision{$74.75$} &
      $598$ &  \textbf{19}\\
    \arrayrulecolor{black}
    \hline
    \end{tabular}
    \label{tab_performanceComparison}
}
\end{table}
\vspace{-0.5ex}

\subsection{Quantitative performance}
\label{sec_timePerformance}

The time complexity of each iteration
of the baseline optimization is cubic in the worst case, but quadratic in
practice (\autoref{sec_approach}). As discussed in \autoref{sec_algorithm}, our
approach has the same worst case complexity, but behaves more efficiently
in practice thanks to our  accelerations.

\autoref{tab_performanceComparison}
provides an overall comparison between the baseline optimization
(\autoref{sec_approach}) and our solver (\autoref{sec_algorithm}).
Specifically, it compares both approaches in terms of
runtime, for
a basic simplification scenario\revision{:} \emph{non-signal}  pairs are
identified as the input pairs with a persistence smaller than $1\%$ of the
function range \revision{(see Appendix A for an aggressive simplification
scenario)}.
For both
approaches, we set the stopping criterion $s$ to $0.01$, such that both methods
reach a similar residual loss at termination (and hence produce results of
comparable quality).
This table shows that for
a basic simplification scenario,
our approach produces results within minutes (at most \discuss{35}).
In contrast, the baseline approach does
not produce a result after 24 hours of computation for the largest
examples.
Otherwise, it still exceeds hours of computation
for diagrams of modest size.
Overall, our approach results in an average \discuss{$\times 64$} speedup.
This
acceleration can be explained by several factors. First, the direct gradient
descent (\autoref{sec_gradientDescent}) requires \emph{fewer} iterations than
the
baseline approach (we discuss this further in the next paragraph, presenting
\autoref{tab_qualityComparison}). Second, our approach also results in
\emph{faster} iterations, given
the accelerations presented in
\autoref{sec_algorithm}.

\revision{In practice, the overall runtime for our solver is a function of the
size of the input and target diagrams (large diagrams lead to large
assignment problems).
The size of the topological features in the geometric domain also plays a role
(larger features will require more iterations). Finally, the number of still
\emph{signal} pairs also plays a role given our fast assignment update procedure
(\autoref{sec_fastDistance}, a large number of still \emph{signal} pairs leads
to faster assignments).
For instance, the total number of pairs (input plus target) for the
\emph{Neocortical Layer Axon} dataset is about $\times20$ larger than
that of
the \emph{Aneursym} dataset, and the ratio between their respective runtime is
also about $20$. Moreover, the \emph{Foot} and \emph{Neocortical Layer Axon}
datasets have comparable overall sizes.
However, the latter
dataset results in a computation time
$\times5$ larger.
This can be partly
explained by the fact that the topological features are larger in this
dataset, yielding twice more iterations (hence explaining a $\times2$
slowdown). Moreover, the per-iteration runtime is also $\times2.5$
slower
(explaining the overall $\times5$ slowdown), due the higher
percentage of non-still \emph{signal} pairs,
increasing
the size of the
assignment problem.}

The runtime gains provided by
\revision{our individual}
accelerations \revision{are} presented in
\autoref{tab_gains}.
\revision{Specifically,}
our procedure for fast Persistence update (\autoref{sec_fastPersistenceUpdate})
can save up to \discuss{$41.4\%$} of overall computation time\revision{, and
$6.6\%$ on average}.
Our
procedure for
fast assignment update (\autoref{sec_fastDistance}) provides the most
substantial gains, saving up to \discuss{$97\%$} of the overall computation
time for the largest target diagram\revision{, and $76\%$ on average (see
Appendix A for a discussion regarding aggressive simplifications)}.

\begin{table}
\vspace{-1.5ex}
 \caption{Individual gains (in
percentage of runtime) for each of our accelerations for
the topological simplification parameters used in
\autoref{tab_performanceComparison}.\vspace{-1ex}}
    \label{tab_gains}
    \fontsize{7}{7}\selectfont
    \def\arraystretch{0.5}
    \setlength{\tabcolsep}{0.57em}
    \adjustbox{width=\linewidth,center}{
        \begin{tabular}{|l|r|r|r||r||r|}
        \hline
        \multirow{2}{*}{Dataset}&
        \multirow{2}{*}{$d$}    &
        \multirow{2}{*}{$|\diagram(f)|$} &
        \multirow{2}{*}{$|\target|$} &
        Persistence Update  &
        Assignment Update  \\
        &
        &
        &
        &
        \multicolumn{1}{c||}{(\autoref{sec_fastPersistenceUpdate})} &
        \multicolumn{1}{c|}{(\autoref{sec_fastDistance})} \\
        \hhline{|=|=|=|=||=||=|}
        Cell &
        $2$ &
        $7,676$ &
      $2,635$ &
        $5.6$ &
        $79.2$\\
        \arrayrulecolor{lightgray}
        \hline
        Ocean Vortices &
        $2$ &
        $12,069$ &
      $2,781$ &
        $-0.4$ &
        $79.8$\\
        \arrayrulecolor{black}
        \hline
        Aneurysm &
        $3$ &
        $38,490$ &
      $24,725$ &
        $41.4$ &
        $24.8$\\
        \arrayrulecolor{lightgray}
        \hline
        Bonsai &
        $3$ &
        $168,489$ &
      $55,464$ &
        $12.1$ &
        $80.6$\\
        \arrayrulecolor{lightgray}
        \hline
        Foot &
        $3$ &
        $754,965$ &
      $474,271$ &
        $0.2$ &
        $90.9$\\
        \arrayrulecolor{lightgray}
        \hline
        Neocortical Layer Axon &
        $3$ &
        $765,406$ &
      $483,791$ &
        $0.0$ &
        $74.6$\\
        \arrayrulecolor{lightgray}
        \hline
        Dark Sky &
        $3$ &
        $1,140,653$ &
      $774,793$ &
        $3.2$ &
        $96.7$\\
        \arrayrulecolor{lightgray}
        \hline
        Backpack &
        $3$ &
        $1,331,362$ &
      $84,402$ &
        $1.1$&
        $74.5$\\
        \arrayrulecolor{lightgray}
        \hline
        Head Aneurysm &
        $3$ &
        $1,345,168$ &
      $234,672$ &
        $1.3$ &
        $93.4$\\
        \arrayrulecolor{lightgray}
        \hline
        Chameleon &
        $3$ &
        $3,641,961$ &
      $32,578$ &
        $1.5$ &
        $61.3$\\
        \arrayrulecolor{black}
        \hline
        \end{tabular}
    }
\end{table}

\begin{table}
\caption{Quality comparison between the baseline optimization
approach (\autoref{sec_approach}) and our solver (\autoref{sec_algorithm}) for
the
parameters used in
\autoref{tab_performanceComparison}.\vspace{-1ex}}
    \fontsize{7}{7}\selectfont
    \def\arraystretch{0.5}
    \setlength{\tabcolsep}{0.57em}
\adjustbox{width=\linewidth,center}{
    \begin{tabular}{|l|r||r|r|r||r|r|r|}
    \hline
    \multirow{2}{*}{Dataset}
    & \multirow{2}{*}{$d$} &
    \multicolumn{3}{c||}{Baseline (\autoref{sec_approach})} &
    \multicolumn{3}{c|}{Our solver (\autoref{sec_algorithm})} \\
    ~
    & ~ &
        $\loss(\dataVector_g)$
        &
        $||f-g||_2$ &
        $||f-g||_\infty$ &
         $\loss(\dataVector_g)$
          &
          $||f-g||_2$ &
        $||f-g||_\infty$
        \\
    \hhline{|=|=||=|=|=||=|=|=|}
      Cells & $2$ &
      $0.0003$ &
      $0.2939$ &
      $0.0054$ &
      $0.0003$&
      $0.2996$ &
      $0.0070$\\
      \arrayrulecolor{lightgray}
      \hline
      Ocean Vortices & $2$ &
        $0.0006$ &
        $0.3710$ &
        $0.0055$ &
        $0.0005$ &
        $0.3769$ &
        $0.0074$\\
       \arrayrulecolor{black}
      \hline
      Aneurysm & $3$ &
      $0.0007$ &
      $0.3481$ &
      $0.0063$ &
      $0.0006$&
      $0.3174$&
      $0.0117$\\
       \arrayrulecolor{lightgray}
      \hline
      Bonsai & $3$ &
        $0.0064$ &
        $1.0243$ &
        $0.0060$ &
        $0.0053$ &
        $1.0541$&
        $0.0117$\\
       \arrayrulecolor{lightgray}
      \hline
      Foot & $3$ &
        $0.0326$&
        $2.0526$ &
        $0.0058$ &
      $0.0297$&
      $2.0483$&
      $0.0127$\\
       \arrayrulecolor{lightgray}
      \hline
      Neocortical Layer Axon & $3$ &
        $0.0271$ &
      $2.0876$ &
      $0.0085$ &
    $0.0279$ &
      $2.1435$ &
      $0.0154$ \\
      \arrayrulecolor{lightgray}
      \hline
      Dark Sky & $3$ &
      NA &
      NA &
      NA &
      $0.0166$ &
      $1.8617$ &
      $0.0148$ \\
      \arrayrulecolor{lightgray}
      \hline
      Backpack & $3$ &
      $0.0339$ &
      $2.4438$ &
      $0.0070$ &
      $0.0312$&
      $2.1991$&
      $0.0159$\\
      \arrayrulecolor{lightgray}
      \hline
      Head Aneurysm & $3$ &
      NA &
      NA &
      NA &
      $0.0750$ &
      $3.7571$ &
      $0.0130$\\
      \arrayrulecolor{lightgray}
      \hline
      Chameleon & $3$ &
      $0.1679$ &
      $5.1264$ &
      $0.0055$ &
      $0.1736$ &
      $4.7773$&
      $0.0143$\\
    \arrayrulecolor{black}
    \hline
    \end{tabular}
    \label{tab_qualityComparison}
}
\end{table}

\begin{figure*}
    \vspace{-1ex}
    \centering
    \includegraphics[width=\linewidth]{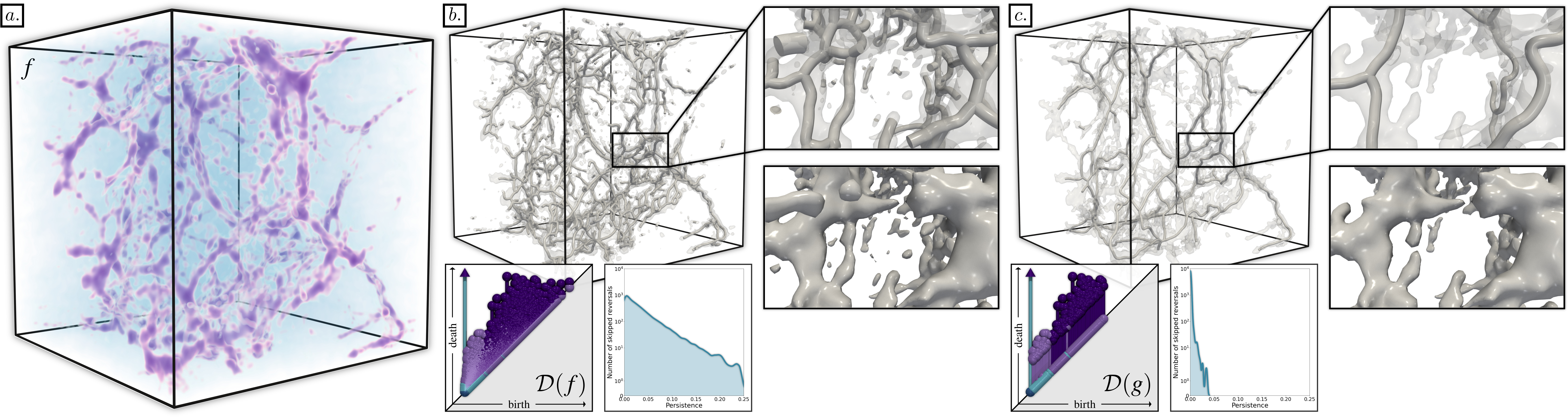}
    \vspace{-2ex}
    \caption{Topological simplification optimization
    for a
challenging dataset
(\emph{``Dark Sky''}: dark matter
density in a cosmology simulation, $(a)$, \emph{signal pairs}: pairs with
a persistence larger than $0.25$).
The
geometry of the \emph{cosmic web} \cite{sousbie11, shivashankar2016felix} is
captured $(b)$ by an isosurface (at isovalue $0.4$) and its core filament
structure is extracted by
the
upward
discrete integral lines, started at $2$-saddles above $0.4$.
The latter
structure contains many small-scale loops as many,
persistent saddle connector reversals
could not be performed
(bottom left histogram).
The local minimum $g$  of the simplification energy (\autoref{eq_energy})
found by our solver $(c)$ has a number of
\emph{non-signal} pairs
reduced by \discuss{$92\%$}.
This
results in a
less cluttered visualization,
as the cosmic web has a
less complicated topology (noisy connected components are
removed and small scale handles are cut, inset zooms). This also induces
fewer skips of persistent saddle connector reversals
(bottom right histogram), hence simplifying more
loops and revealing the main filament structure.
    }
    \label{fig_darkSky}
    \vspace{-2ex}
\end{figure*}

\begin{figure}
    \includegraphics[width=\linewidth]{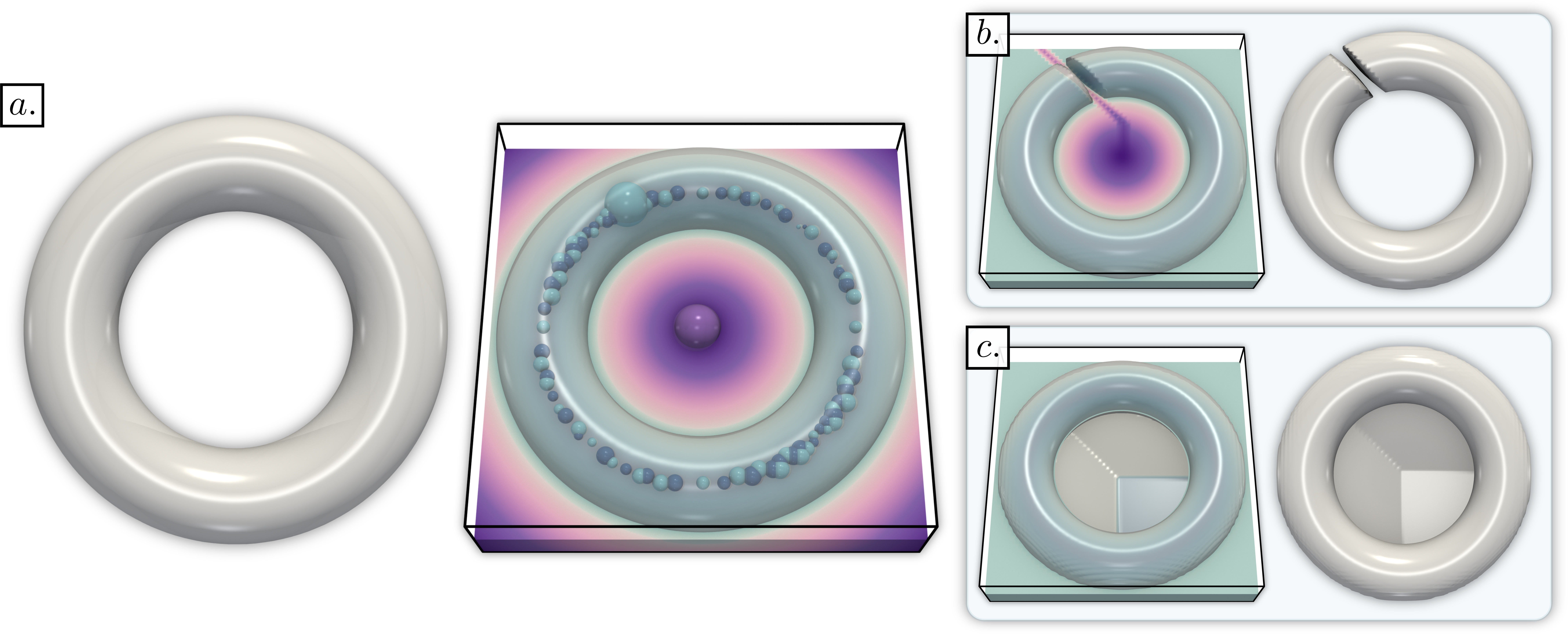}
    \caption{Handle removal on a torus example. $(a)$ The input surface
$\surface$ (left) is used to compute a 3D signed distance field $f$
(right, color map).
$f$
contains a persistent saddle pair (large
spheres) encoding the handle of the torus and many low-persistence
minimum-saddle pairs (smaller spheres, radius scaled by persistence) which are
artifacts (located on the medial axis of $\surface$) of the sampling of the
distance field (which has discontinuous derivatives). The handle can be removed
in the output surface $\surface'$ $(b, c)$ by considering the zero level set of
a simplified field $g$ obtained with our approach (in this example, only the
persistent generator of $f$ with infinite persistence has been maintained).
The handle can be removed either by \emph{cutting} $(b)$ (by only using the
\emph{birth gradient}, \autoref{eq_birthAndDeathGradients}) or by \emph{filling}
$(c)$ (by only using the \emph{death gradient},
\autoref{eq_birthAndDeathGradients}).\vspace{-1ex}}
    \label{fig_torusRepair}
\end{figure}

\autoref{tab_qualityComparison} compares the quality of the output obtained
with the baseline optimization (\autoref{sec_approach}) and our algorithm
(\autoref{sec_algorithm}), for the simplification parameters used
in \autoref{tab_performanceComparison}. The quality is estimated based on the
value of the loss at termination ($\loss(\dataVector_g)$), which assesses the
quality of the topological simplification.
To estimate the proximity of the solution $g$ to the input $f$, we also
evaluate
the
distances $||f-g||_2$ (giving a global error for the entire dataset) and
$||f-g||_\infty$ (giving a pointwise worst case error). 
\revision{We refer the reader to Appendix B for complementary quality
statistics.}
Overall, 
\revision{\autoref{tab_qualityComparison}}
shows that our approach provides comparable losses to the baseline approach
(sometimes marginally better). In terms of data fitting, our approach also
provides comparable global distances $||f-g||_2$ (sometimes marginally better).
For the pointwise worst case error ($||f-g||_\infty$), our approach
\revision{can result}
in
degraded values (by
a factor $2$). This can be explained by
the fact that, when tuning the parameters of our approach, we optimized
the gradient step size to minimize running time, hence possibly triggering in
practice bigger pointwise shifts in data values.
In contrast, the baseline approach uses the Adam \cite{KingmaB14} algorithm,
which optimizes step sizes along the iterations, possibly triggering milder
pointwise shifts in data values.
In principle, the
$||f-g||_\infty$ distance could be improved for our solver by considering
smaller step sizes, but at the expense of more iterations.

\subsection{Analyzing topologically simplified data}
\label{sec_dataSimplification}
Our approach enables the direct visualization and analysis of topologically
simplified data. This is illustrated in \autoref{fig_teaser}, which shows the
processing of an acquired dataset (\emph{``Aneurysm''}) representing a network
of arteries.
As documented in the literature \cite{MarinoK16, HuBMK23}, this network
exhibits a typical tree-like structure, whose accurate
\revision{geometric}
extraction is
relevant for medical analysis. The filament structure of the arteries can be
simply extracted
by considering the
\emph{discrete integral lines}
\cite{guillou_tvcg23} (a.k.a. \emph{v-paths} \cite{forman98}) which connect
$2$-saddles to maxima and which have a minimum function value above $0.1$
(scalar fields are normalized). This value $0.1$ generates an
isosurface (transparent surfaces, \autoref{fig_teaser}) which accurately
captures the geometry of the blood vessels. Hence, selecting the discrete
integral lines above that threshold guarantees the extraction of the filament
structures within the vessels.

As shown in \autoref{fig_teaser}, the
diagram $\diagram(f)$ contains several saddle pairs, corresponding to
persistent $1$-dimensional generators \cite{guillou_tvcg23, iuricich21} (curves
colored by persistence in the inset zooms), which yields incorrect loops in the
filament structure (which is supposed to have a tree-like structure
\cite{HuBMK23}). To remove loops in networks of discrete integral
lines,
an established
topological technique,
relying on standard discrete Morse theory \cite{forman98},
consists in reversing the discrete gradient
\cite{GuntherRSW14} along \emph{saddle connectors}.
We recap this procedure here for completeness. Given the persistence diagram
$\diagram(f)$, we process its
\emph{non-signal}
saddle pairs in increasing order of persistence.
For each saddle pair $(\simplex_b,
\simplex_d)$,
its \emph{saddle connector} is constructed
by following the discrete gradient of
$f$ from $\simplex_d$ down to $\simplex_b$. Next, the pair of critical
simplices $(\simplex_b,
\simplex_d)$ is cancelled, in the discrete sense, by simply reversing the
discrete gradient along its saddle connector \cite{forman98}  (i.e., each
discrete vector is reversed to point to the
preceding co-face).
Such a reversal is marked as \emph{valid} if it does not
create any cycle in the discrete gradient field. The validity of a reversal is
important since invalid reversals result in
discrete vector fields which no longer describe valid scalar fields, and
from which the subsequent extraction of integral lines can generate further
loops (which we precisely aim to remove).
The cancellation of a saddle pair $(\simplex_b, \simplex_d)$ is \emph{skipped}
if the reversal of its saddle connector is not valid, or if its saddle
connector does not exist. The latter
case
occurs for instance for
nested saddle pairs, when an invalid reversal of a small persistence pair
prevents the subsequent reversal of a larger
one.
Finally, when all
the \emph{non-signal} saddle pairs
have been
processed,
the simplified filament
structures are simply
obtained
from
the simplified discrete gradient,
by
initiating
integral lines from $2$-saddles up to maxima.

However, in the example of \autoref{fig_teaser}, this saddle connector
reversal procedure fails at simplifying the spurious loops in the filament
structures, while maintaining a valid discrete gradient
(\autoref{fig_teaser}$(b)$).
As discussed in the literature \cite{GuntherRSW14},
integral line reversal is
indeed not guaranteed to completely simplify saddle pairs
\revision{(\emph{v-path} co-location \cite{IuricichFF15} as well as
specific cancellation orderings \cite{GyulassyNPBH05, gyulassy_vis07}
can challenge reversals,
the latter issue being a manifestation of
the NP-hardness of the problem
\cite{AttaliBDGL13}).}
This is evaluated in
the bottom left histogram, which reports the number of \emph{skipped} saddle
connector reversals as a function of the persistence of the corresponding pair.
Specifically, this histogram shows that the reversal of several high-persistence
saddle pairs could not be performed, hence the presence of
large
loops in
the extracted filament structures.

\begin{figure*}
    \vspace{-1ex}
    \centering
    \includegraphics[width=\linewidth]{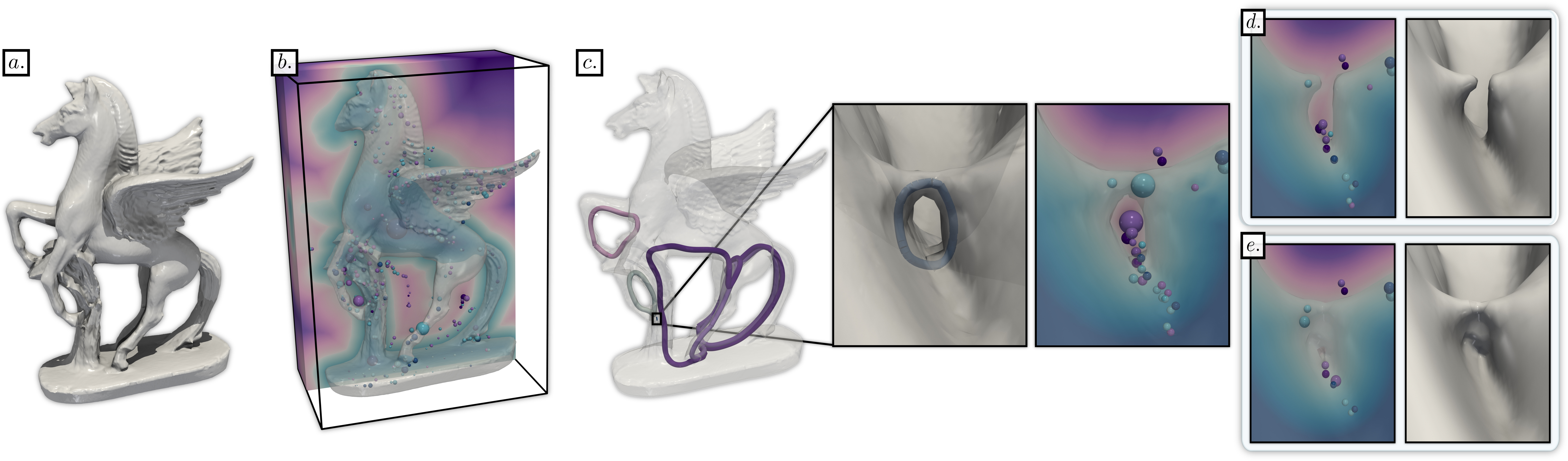}
    \vspace{-4ex}
    \caption{Removal of a spurious handle from an acquired
surface $\surface$
$(a)$. First, the signed distance field $f$ is computed from $\surface$ $(b)$.
$f$ is shown with a color map on the clipped volume, with its critical
simplices colored per dimension, with a sphere with a radius proportional to
their persistence. The extraction of the $1$-dimensional persistent generators
\cite{guillou_tvcg23, iuricich21} ($(c)$, colored by persistence)
reveals the existence of a short generator in $f$, corresponding to
a small handle defect in $S$
(under the Pegasus front left hoof, see inset zooms).
Our framework can repair this defect by simplifying the
corresponding
saddle
pair,
either by \emph{cutting} ($(d)$, by only using the \emph{birth} gradient,
\autoref{eq_birthAndDeathGradients}) or by \emph{filling} ($(e)$, by only
using the \emph{death} gradient,
\autoref{eq_birthAndDeathGradients}).
\vspace{-2ex}}
    \label{fig_pegasus}
\end{figure*}

Our approach can be used to efficiently generate a function $g$ which is close
to the input $f$ and from which the removal of saddle pairs has been optimized,
while maintaining intact the rest of the features (see the resulting diagram
$\diagram(g)$, \autoref{fig_teaser}). Specifically,
we set as \emph{non-signal} pairs \emph{all} the saddle pairs of the input, and
we
set as \emph{signal} pairs all the others (irrespective of their persistence).
This
enables a direct visualization and analysis of the topologically
simplified data, where isosurface handles have been cut
(\autoref{fig_teaser}\revision{c}, bottom\revision{-right} zoom \revision{vs.
\autoref{fig_teaser}\revision{b}, bottom\revision{-right} zoom}) and where
most spurious filament loops have been
consequently
simplified
(\autoref{fig_teaser}, top zoom).
Note that, as shown in the bottom right histogram, our optimization modifies
the input data $f$ into a function $g$ where reversal skips still occur.
This is due to the fact that our solver  identifies a \emph{local} minimum
of the simplification energy (\autoref{eq_energy}) and that, consequently, a
few saddle pairs,
with  low persistence,
may still remain (we recall that
\revision{sublevel set simplification}
is NP-hard \cite{AttaliBDGL13}, see \autoref{sec_limitations} for
further discussions). However,
the skipped reversals  which remain after our optimization
(\autoref{fig_teaser}, bottom right histogram)
only involve very low persistence pairs, hence
allowing the cancellation of the largest loops overall.

\autoref{fig_darkSky} illustrates our
simplification optimization
for a challenging dataset (\emph{``Dark Sky''}: dark matter density in a
cosmology simulation).
\revision{The}
isosurface capturing the \emph{cosmic web}
\cite{sousbie11, shivashankar2016felix} \revision{(inset zooms)} has a
complicated topology (many noisy
connected components and handles), which challenges its visual inspection. Its
core filament structure also contains many small-scale loops since many
persistent saddle connector reversals could not be performed
(\autoref{fig_darkSky}, bottom left
histogram). Our solver provides a local minimum $g$ to the simplification energy
(\autoref{eq_energy})
with a number of \emph{non-signal} pairs
reduced by \discuss{$92\%$} \revision{(see Appendix C for further stress
experiments)}.
This results in a less cluttered visualization, as the resulting
cosmic web (\autoref{fig_darkSky}$(c)$) has a less complicated topology (noisy
connected
components are removed and small scale handles are cut, inset zooms). Moreover,
our optimization modifies the data in a way that is more conducive to
persistent
saddle connector reversals (bottom right histogram),
hence simplifying more loops and, thus, better revealing overall the large-scale
filament structure of the cosmic web.

\subsection{Repairing genus defects in surface processing}
\label{sec_genusReparation}

Our work can also be used to repair genus defects in surface
processing, where surface models, in particular when they are acquired, can
include spurious handles due to acquisition artifacts.
\revision{While several approaches have been proposed to address this issue
\cite{ChambersJLLTY18, ZengCLJ20, ZengCLJ22}, they typically rely on intensive
automatic optimizations, aiming at selecting 
the \emph{best} sequence of
local simplification primitives (i.e. \emph{cutting} or \emph{filling}). In
contrast, our approach relies on a simpler and lightweight procedure, which
provides control to the user over the primitives to use.
Moreover, most existing techniques simplify only one sublevel set, while our
approach processes the whole function range.}
For this, we
consider
the three-dimensional signed distance field \revision{$f$} to
the input surface $\surface$, computed on a regular grid (i.e., $f$ encodes for
each grid vertex $v$ the distance to the closest point on the
surface \revision{$\surface$}, multiplied by $-1$ if $v$ is located within the
volume enclosed by
$\surface$). For such a field, the
zero level set $f^{-1}(0)$ coincides with
$\surface$. Then, the removal of a handle in $\surface$ can be performed by
creating a simplified signed distance field $g$, where the corresponding saddle
pair has been canceled. Finally, the zero level set $g^{-1}(0)$ provides the
simplified surface $\surface'$.

This process is illustrated in
\autoref{fig_torusRepair} where the handle of a torus is removed. Note that,
from a topological point of view, this operation can be performed in two ways:
either by cutting the handle (\autoref{fig_torusRepair}$(b)$), or by filling it
(\autoref{fig_torusRepair}$(c)$).
This can be
controlled in our solver by
simply
adjusting
the
step sizes for the birth and death gradients
(\autoref{sec_gradientDescent}). Specifically,
given a saddle pair to remove  $p_i \in \diagram(f)$,
handle cutting is
obtained by setting
$\alpha_d$
to zero. Then,
the  death vertex $v_{i_d}$
will
not be modified
(above the zero level set), while only the
birth vertex $v_{i_b}$ (located in the star of the $1$-saddle
creating
the handle) will increase its value above $0$, effectively
disconnecting the handle in the output surface $\surface'$. Handle filling is
obtained symmetrically, by setting
$\alpha_b$ to
zero (effectively forcing the $2$-saddle to decrease its value below $0$).

\autoref{fig_pegasus} presents a realistic example of an acquired surface
from a public repository \cite{ttkData},
which
contains a spurious handle, due to acquisition artifacts. First, the signed
distance field
is computed and its $1$-dimensional persistent generators
\cite{iuricich21, guillou_tvcg23} are extracted. The shortest generator
corresponds to a small handle, which happens to be a genus defect in this
example. Then, the user can choose to repair this defect via
cutting or filling, resulting in a
repaired surface $S'$ which is close to the input $S$, and from which the
spurious handle has been removed.

\subsection{Limitations}
\label{sec_limitations}
Our approach is
essentially numerical and, thus, suffers from the same limitations as previous
numerical methods for topological simplification (\autoref{sec_relatedWork}).
Specifically,
the \emph{non-signal} pairs
are canceled by our approach by decreasing
their persistence to a target value of zero. However, this
decrease
is
ultimately
limited by the employed numerical precision (typically, $10^{-6}$ for
single-precision floating point values).
From a strictly combinatorial point of
view, this can result in \emph{residual pairs} with an arbitrarily small
persistence (i.e., in the order of the numerical precision).
In principle, this drawback is common
to all numerical methods (although sometimes mitigated via smoothing).
Then, when computing topological abstractions,
these residual pairs need to be removed from
the computed abstraction (e.g., with integral line reversal,
\autoref{sec_dataSimplification}). However, as discussed in the literature
\cite{GyulassyBHP11, GuntherRSW14}, post-process mechanisms for simplifying
topological abstractions may not guarantee a complete simplification of the
abstractions either (this is another concrete implication of the NP-hardness
of
\revision{sublevel set simplification}
\cite{AttaliBDGL13}). However, our experiments
(\autoref{sec_dataSimplification}) showed that our numerical optimization
helped such combinatorial mechanisms,  by pre-processing the data in a
way that resulted eventually in fewer persistent reversal skips
(Figs. \ref{fig_teaser} and \ref{fig_darkSky},
right versus left histograms).

Similar to previous persistence optimization frameworks, our
 approach generates a \emph{local} minimum of the simplification energy
(\autoref{eq_energy}), and thus it is not guaranteed to reach the global
minimum.
As a reminder,
in 3D,
\revision{an optimal simplification}
(i.e., $\diagram(g) = \target$)
may not exist
and finding
\revision{a sublevel set simplification is}
NP-hard \cite{AttaliBDGL13}.
However, our experiments (\autoref{sec_timePerformance})
showed that our approach still generated solutions
whose quality was on par with the state-of-the-art
(comparable losses and distances to the input),
while providing
substantial accelerations. Moreover, as shown  in
\autoref{sec_dataSimplification}, these solutions enabled the direct
visualization of isosurfaces whose topology was indeed simplified (fewer
 components and handles) and they were also conducive to improved
saddle connector reversals.

%% file: conclusion.tex
\section{Conclusion}
\label{sec_conclusion}

This paper introduced a practical solver for topological
simplification optimization. Our solver is based on tailored accelerations,
which are specific to the problem of topological simplification.
\discuss{Our accelerations are simple and easy to implement}, but result in
significant
gains in terms of runtime, with
\discuss{$\times 60$} speedups
on average on our datasets
over
state-of-the-art
persistence optimization frameworks (with both fewer and faster iterations),
for comparable output qualities.
This makes
topological
simplification optimization practical for real-life
three-dimensional
datasets.
We showed that our contributions enabled a direct visualization and
analysis of
the topologically simplify data, where the topology of the extracted
isosurfaces was indeed simplified (fewer connected components and handles).
We applied our
approach
to the extraction of prominent
filament structures in
3D
data, and showed that our
pre-simplification of the data led to practical improvements for the removal of
spurious loops in filament structures.
We showed that our contributions could be used to repair
genus defects in surface processing, where handles due to acquisition
artifacts could be easily removed,
with an explicit control on the repair primitives (cutting or
filling).

While it is tailored to the problem of simplification, our solver
is still generic and could in principle be used for other persistence
optimization problems,
however, with possibly less important
performance gains. In
the future, we will consider other optimization problems and investigate
other
acceleration strategies for these specific problems.
\revision{Since our solver can optimize persistence pairs
localized within a neighborhood of the field, we will also investigate
divide-and-conquer parallelizations.}

%% file: appendixBody.tex
\appendix

\input{appendixAggressiveSimplification.tex}

\input{appendixSignalPairs.tex}

\input{appendixFullSimplification.tex}

%% file: appendixAggressiveSimplification.tex
\begin{table}
\caption{\revision{Time performance comparison between the
baseline optimization approach (Section 3, main manuscript) and our solver
(Section 4, main manuscript), for
an aggressive simplification
(\emph{non-signal}
pairs: input pairs less persistent than
$45\%$
of the function range).
\revision{The column N.S.S.P. reports the average percentage of non-still
\emph{signal} pairs (per iteration) for our solver.}
The stopping condition
is set to $s = 0.01$.}}
    \fontsize{7}{7}\selectfont
    \def\arraystretch{0.5}
    \setlength{\tabcolsep}{0.57em}
\adjustbox{width=\linewidth,center}{
    \begin{tabular}{|l|r|r|r||r|r|r||r|r|r|r|r|}
    \hline
    \multirow{2}{*}{\revision{Dataset}}
    & \multirow{2}{*}{\revision{$d$}}
    & \multirow{2}{*}{\revision{$|\diagram(f)|$}}
    & \multirow{2}{*}{\revision{$|\target|$}} &
    \multicolumn{3}{c||}{\revision{Baseline (Section 3)}} &
    \multicolumn{5}{c|}{\revision{Our solver (Section 4)}} \\
    ~
    & ~
    & ~
    & ~ &
        \revision{\#It.}
        &
        \revision{Time/It (s.)}
        &
        \revision{Time (s.)} &
        \revision{N.S.S.P.} &
        \#It.
          &
          \revision{Time/It (s.)}
          &
        \revision{Time (s.)}
        & \revision{Speedup}
        \\
    \hhline{|=|=|=|=||=|=|=||=|=|=|=|=|}
       \revision{Cells}  &
       \revision{2} &
       \revision{$7,676$} &
       \revision{$21$} &
       \revision{$755$} &
       \revision{\revision{$0.26$}} &
       \revision{$193$} &
       \revision{\revision{$0.00\%$}} &
       \revision{$167$}&
       \revision{\revision{$0.19$}} &
       \revision{$32$}&
       \revision{\textbf{6}}\\
      \arrayrulecolor{lightgray}
      \hline
       \revision{Ocean Vortices} &
       \revision{$2$} &
       \revision{$12,069$} &
       \revision{$80$} &
       \revision{$474$} &
       \revision{\revision{$0.26$}} &
       \revision{$126$} &
       \revision{\revision{$0.70\%$}} &
       \revision{$269$} &
       \revision{\revision{$0.14$}} &
       \revision{$38$} &
       \revision{\textbf{8}}\\
     \arrayrulecolor{black}
      \hline
       \revision{Aneurysm} &
       \revision{$3$} &
       \revision{$38,490$} &
       \revision{$231$} &
       \revision{$329$} &
       \revision{\revision{$32.65$}} &
       \revision{$10,743$} &
       \revision{\revision{$1.38\%$}} &
       \revision{$17$} &
       \revision{\revision{$8.73$}} &
       \revision{$148$} &
       \revision{\textbf{72}}\\
      \arrayrulecolor{lightgray}
      \hline
       \revision{Bonsai} &
       \revision{$3$} &
       \revision{$168,489$} &
       \revision{$2$} &
       \revision{$483$} &
       \revision{\revision{$72.47$}} &
       \revision{$35,002$} &
       \revision{\revision{$49.23\%$}} &
       \revision{$65$} &
       \revision{\revision{$12.26$}} &
       \revision{$797$} &
       \revision{\textbf{44}}\\
      \arrayrulecolor{lightgray}
      \hline
       \revision{Foot} &
       \revision{$3$} &
       \revision{$754,965$} &
       \revision{$18$} &
       \revision{$108$} &
       \revision{\revision{$36.52$}} &
       \revision{$3,944$} &
       \revision{\revision{$5.05\%$}} &
       \revision{$11$} &
       \revision{\revision{$29.22$}} &
       \revision{$321$} &
       \revision{\textbf{{12}}}\\
      \arrayrulecolor{lightgray}
      \hline
       \revision{Neocortical Layer Axon} &
       \revision{$3$} &
       \revision{$765,406$} &
       \revision{$174$} &
       \revision{$177$} &
       \revision{\revision{$12.54$}} &
       \revision{$2,219$} &
       \revision{\revision{$1.87\%$}} &
       \revision{$8$} &
       \revision{\revision{$15.08$}} &
       \revision{$121$} &
       \revision{\textbf{18}}\\
      \arrayrulecolor{lightgray}
      \hline
       \revision{Dark Sky} &
       \revision{$3$} &
       \revision{$1,140,653$} &
       \revision{$120,332$} &
       \revision{NA} &
       \revision{\revision{NA}} &
       \revision{\textbf{> 24h}} &
       \revision{\revision{$0.38\%$}} &
       \revision{$7$} &
       \revision{\revision{$30.73$}} &
       \revision{$215$} &
       \revision{\textbf{> 402}}\\
      \arrayrulecolor{lightgray}
      \hline
       \revision{Backpack} &
       \revision{$3$} &
       \revision{$1,331,362$} &
       \revision{$97$} &
       \revision{$329$} &
       \revision{\revision{$40.36$}} &
       \revision{$13,278$} &
       \revision{\revision{$2.21\%$}} &
       \revision{$28$} &
       \revision{\revision{$23.05$}} &
       \revision{$646$} &
       \revision{\textbf{21}}\\
      \arrayrulecolor{lightgray}
      \hline
       \revision{Head Aneurysm} &
       \revision{$3$} &
       \revision{$1,345,168$} &
       \revision{$2$} &
       \revision{$97$} &
       \revision{\revision{$119.90$}} &
       \revision{$11,630$} &
       \revision{\revision{$5.26\%$}} &
       \revision{$19$} &
       \revision{\revision{$35.99$}} &
       \revision{$684$} &
       \revision{\textbf{17}}\\
      \arrayrulecolor{lightgray}
      \hline
       \revision{Chameleon} &
       \revision{$3$} &
       \revision{$3,641,961$} &
       \revision{$2$} &
       \revision{NA} &
       \revision{NA} &
       \revision{\textbf{> 24h}} &
       \revision{\revision{$0.00\%$}} &
       \revision{$81$} &
       \revision{\revision{$28.70$}} &
       \revision{$2,325$} &
       \revision{\textbf{> 37}}\\
    \arrayrulecolor{black}
    \hline
    \end{tabular}
    \label{tab_performanceComparisonAppendix}
}
\end{table}

\begin{table}
 \caption{\revision{Individual gains (in
percentage of runtime) for each of our accelerations for
the topological simplification parameters used in
\autoref{tab_performanceComparisonAppendix}.}}
    \label{tab_gainsAppendix}
    \fontsize{7}{7}\selectfont
    \def\arraystretch{0.5}
    \setlength{\tabcolsep}{0.57em}
    \adjustbox{width=\linewidth,center}{
        \begin{tabular}{|l|r|r|r||r||r|}
        \hline
        \multirow{2}{*}{\revision{Dataset}}&
        \multirow{2}{*}{\revision{$d$}}    &
        \multirow{2}{*}{\revision{$|\diagram(f)|$}} &
        \multirow{2}{*}{\revision{$|\target|$}} &
        \revision{Persistence Update}  &
        \revision{Assignment Update}  \\
        &
        &
        &
        &
        \multicolumn{1}{c||}{\revision{(Section 4.2)}} &
        \multicolumn{1}{c|}{\revision{(Section 4.3)}} \\
        \hhline{|=|=|=|=||=||=|}

        \revision{Cell} &
        \revision{$2$} &
        \revision{$7,676$} &
      \revision{$21$} &
        \revision{$17.9$} &
        \revision{$21.9$}\\
        \arrayrulecolor{lightgray}
        \hline
        \revision{Ocean Vortices} &
        \revision{$2$} &
        \revision{$12,069$} &
      \revision{$80$} &
        \revision{$14.9$} &
        \revision{$41.4$}\\
        \arrayrulecolor{black}
        \hline
        \revision{Aneurysm} &
        \revision{$3$} &
        \revision{$38,490$} &
      \revision{$231$} &
        \revision{$54.9$} &
        \revision{$7.6$}\\
        \arrayrulecolor{lightgray}
        \hline
        \revision{Bonsai} &
        \revision{$3$} &
        \revision{$168,489$} &
      \revision{$2$} &
        \revision{$44.8$} &
        \revision{$1.5$}\\
        \arrayrulecolor{lightgray}
        \hline
        \revision{Foot} &
        \revision{$3$} &
        \revision{$754,965$} &
      \revision{$18$} &
        \revision{$30.0$} &
        \revision{$-21.5$}\\
        \arrayrulecolor{lightgray}
        \hline
        \revision{Neocortical Layer Axon} &
        \revision{$3$} &
        \revision{$765,406$} &
      \revision{$174$} &
        \revision{$0.1$} &
        \revision{$-28.9$}\\
        \arrayrulecolor{lightgray}
        \hline
        \revision{Dark Sky} &
        \revision{$3$} &
        \revision{$1,140,653$} &
      \revision{$120,332$} &
        \revision{$-1.3$} &
        \revision{$91.6$}\\
        \arrayrulecolor{lightgray}
        \hline
        \revision{Backpack} &
        \revision{$3$} &
        \revision{$1,331,362$} &
      \revision{$97$} &
        \revision{$17.2$}&
        \revision{$27.9$}\\
        \arrayrulecolor{lightgray}
        \hline
        \revision{Head Aneurysm} &
        \revision{$3$} &
        \revision{$1,345,168$} &
      \revision{$2$} &
        \revision{$8.8$} &
        \revision{$67.0$}\\
        \arrayrulecolor{lightgray}
        \hline
        \revision{Chameleon} &
        \revision{$3$} &
        \revision{$3,641,961$} &
      \revision{$2$} &
        \revision{$9.0$} &
        \revision{$92.8$}\\
        \arrayrulecolor{black}
        \hline
        \end{tabular}
    }
\end{table}

\section{\revision{Aggressive simplification}}
\revision{Section 5.1 (main manuscript) evaluates our approach from a
quantitative point of view, for a simplification scenario where all the pairs
less persistent than $1\%$ of the function range are considered as
\emph{non-signal}. In this appendix, we report the same experiments, but with a
more aggressive threshold ($45\%$ of the function range).}

\revision{Specifically, \autoref{tab_performanceComparisonAppendix}
provides a comparison between the baseline optimization
(Section 3, main manuscript) and our solver (Section 4, main manuscript) in
terms of runtime.
Similarly to the basic simplification scenario (Table 1, main
manuscript),
our solver computes the simplifications within minutes
(at most $39$), for the same average speedup over the baseline ($\times64$).
For both the baseline and our solver, while the number of
iterations
increases in comparison to the basic simplification (since more and larger
features need to be simplified), iterations are significantly faster as the
assignment problems are
much
smaller.}

\revision{The runtime gains provided by
\revision{our individual}
accelerations, for this aggressive simplification scenario, are presented in
\autoref{tab_gainsAppendix}.
\revision{Specifically,}
our procedure for fast Persistence update (Section 4.2, main manuscript)
can save up to $54.9\%$ of overall computation time\revision{, and
$19.6\%$ on average}, which is a substantial improvement over the basic
simplification scenario (Table 2, main manuscript).
As more iterations are required to
simplify persistent features (\autoref{tab_performanceComparisonAppendix}),
less and
less vertices are updated along the iterations (since low-persistence features
are cancelled in the early iterations), hence advantaging our fast
persistence update procedure.
For the fast Assignment update (Section 4.3, main manuscript),
the average gain decreases to $30\%$ (with regard to the basic simplification
scenario, Table 2, main manuscript) since assignment problems become smaller
(and so does their importance in the overall computation).}
\revision{Negative entries in \autoref{tab_gainsAppendix} indicate cases where
the acceleration
actually degrades runtimes. For the fast persistence update, this
happens when the number of updated vertices is so large that their
identification overweights the gradient computation for the non-updated
vertices. Similar remarks can be made for the fast assignment update, where the
identification of the still pairs can penalize runtime for small assignment
problems.}

\revision{Overall, both our accelerations (fast persistence update
and fast assignment update)
improve performance in both simplification
scenarios, with the fast assignment update being more important for mild
simplifications, and the fast persistence update for aggressive ones.}

\revision{\autoref{tab_qualityComparisonAppendix} compares the quality of the
output obtained
with the baseline optimization (Section 3, main manuscript) and our algorithm
(Section 4, main manuscript), for the simplification parameters used
in \autoref{tab_performanceComparisonAppendix}. In particular, this table
provides similar observations to the basic simplification scenario (Table 3,
main manuscript): our approach provides comparable losses to the baseline
approach
(sometimes marginally better). In terms of data fitting, our approach also
provides comparable global distances $||f-g||_2$ (sometimes marginally better).
For the pointwise worst case error ($||f-g||_\infty$), similarly to the basic
simplification scenario,
our approach can result in degraded values (roughly by a factor of $2$), as
discussed in further details in the main manuscript.}

\begin{table}
\caption{\revision{Quality comparison between the baseline optimization
approach (Section 3, main manuscript) and our solver (Section 4, main
manuscript) for
the
parameters used in
\autoref{tab_performanceComparisonAppendix}.}}
    \fontsize{7}{7}\selectfont
    \def\arraystretch{0.5}
    \setlength{\tabcolsep}{0.57em}
\adjustbox{width=\linewidth,center}{
    \begin{tabular}{|l|r||r|r|r||r|r|r|}
    \hline
    \multirow{2}{*}{\revision{Dataset}}
    & \multirow{2}{*}{\revision{$d$}} &
    \multicolumn{3}{c||}{\revision{Baseline (Section 3)}} &
    \multicolumn{3}{c|}{\revision{Our solver (Section 4)}} \\
    ~
    & ~ &
        $\loss(\dataVector_g)$
        &
        $||f-g||_2$ &
        $||f-g||_\infty$ &
         $\loss(\dataVector_g)$
          &
          $||f-g||_2$ &
        $||f-g||_\infty$
        \\
    \hhline{|=|=||=|=|=||=|=|=|}
     \revision{Cells} &
     \revision{$2$} &
       \revision{$0.0627$} &
       \revision{$10.0828$} &
       \revision{$0.1991$} &
       \revision{$0.0628$}&
       \revision{$9.8787$} &
       \revision{$0.2397$}\\
      \arrayrulecolor{lightgray}
      \hline
       \revision{Ocean Vortices} &
       \revision{$2$} &
         \revision{$0.0547$} &
         \revision{$4.8040$} &
         \revision{$0.1370$} &
         \revision{$0.0541$} &
         \revision{$12.7392$} &
         \revision{$0.4404$}\\
       \arrayrulecolor{black}
      \hline
       \revision{Aneurysm} &
       \revision{$3$} &
       \revision{$1.1040$} &
       \revision{$23.2228$} &
       \revision{$0.2586$} &
       \revision{$1.0278$}&
       \revision{$14.1333$}&
       \revision{$0.4205$}\\
       \arrayrulecolor{lightgray}
      \hline
       \revision{Bonsai} &
       \revision{$3$} &
         \revision{$0.6679$} &
         \revision{$34.3607$} &
         \revision{$0.2014$} &
         \revision{$0.6651$} &
         \revision{$19.0419$}&
         \revision{$0.3712$}\\
       \arrayrulecolor{lightgray}
      \hline
       \revision{Foot} &
       \revision{$3$} &
         \revision{$3.3992$}&
         \revision{$25.7129$} &
         \revision{$0.0967$} &
       \revision{$2.9832$}&
       \revision{$23.7273$}&
       \revision{$0.2375$}\\
       \arrayrulecolor{lightgray}
      \hline
       \revision{Neocortical Layer Axon} &
       \revision{$3$} &
         \revision{$5.5482$} &
       \revision{$29.2017$} &
       \revision{$0.1602$} &
     \revision{$4.6222$} &
       \revision{$28.1093$} &
       \revision{$0.3424$} \\
      \arrayrulecolor{lightgray}
      \hline
       \revision{Dark Sky} &
       \revision{$3$} &
       \revision{NA} &
       \revision{NA} &
       \revision{NA} &
       \revision{$87.4867$} &
       \revision{$116.2727$} &
       \revision{$0.5851$} \\
      \arrayrulecolor{lightgray}
      \hline
       \revision{Backpack} &
       \revision{$3$} &
       \revision{$1.0060$} &
       \revision{$23.0536$} &
       \revision{$0.2397$} &
       \revision{$1.1978$}&
       \revision{$13.8104$}&
       \revision{$0.4043$}\\
      \arrayrulecolor{lightgray}
      \hline
       \revision{Head Aneurysm} &
       \revision{$3$} &
       \revision{$0.4880$} &
       \revision{$16.7538$} &
       \revision{$0.0873$} &
       \revision{$0.4912$} &
       \revision{$10.3221$} &
       \revision{$0.2386$}\\
      \arrayrulecolor{lightgray}
      \hline
       \revision{Chameleon} &
       \revision{$3$} &
       \revision{NA} &
       \revision{NA} &
       \revision{NA} &
       \revision{$0.4223$} &
       \revision{$10.8036$}&
       \revision{$0.2949$}\\
    \arrayrulecolor{black}
    \hline
    \end{tabular}
    \label{tab_qualityComparisonAppendix}
}
\end{table}

%% file: appendixSignalPairs.tex
\section{\revision{Signal pair preservation evaluation}}

\revision{Table 3 (main manuscript) provides some quality statistics regarding
the output of our algorithm. Specifically, it details the achieved loss
($\loss(\dataVector_g)$) as well as pointwise distances between the input and
the simplified fields ($||f-g||_2$ and $||f-g||_\infty$).}

\revision{In this appendix, we provide complementary quality statistics, where we now evaluate the preservation of the \emph{features of interest} after our simplification. 
For this, \autoref{tab_signalPairPreservation} reports statistics (minimum,
average, maximum) of the displacement in the birth-death space (between $0$ and
$1$) for the \emph{signal} pairs, both for a mild (white lines) and an
aggressive (grey lines) simplification based on persistence ($1\%$ and $45\%$ of
the function range, respectively).
Specifically, displacements are evaluated given the optimal assignment (achieved
by the Wasserstein distance) between
$\target$
and $\diagram(g)$.
Overall, this table shows that the position of the \emph{signal} pairs in the
birth-death space is well constrained by our solver,
with a worst displacement
of $2.25\times10^{-02}$ for a challenging example (aggressive
simplification of the \emph{Dark Sky} dataset, where many
multi-saddles are involved in both \emph{signal} and \emph{non-signal} pairs).
For all datasets, the achieved worst displacement is
negligible with regard to the employed persistence threshold (by an
order of magnitude).}

\begin{table}
\caption{\revision{Statistics (minimum, average, maximum) of displacement in the
birth-death space (between $0$ and $1$) for the \emph{signal} pairs. The employed
simplification parameters are those used in the Table 1 of the main manuscript
(white lines: mild simplification, grey lines: aggressive one).}}
    \fontsize{7}{7}\selectfont
    \def\arraystretch{0.5}
    \setlength{\tabcolsep}{0.57em}
\adjustbox{width=\linewidth,center}{
    \begin{tabular}{|l|r||r|r|r|}
    \hline
    \revision{Dataset} &
    \revision{$d$} &
    \revision{Min.} &
    \revision{Avg.} & 
    \revision{Max.}\\
    \hhline{|=|=||=|=|=|}
    \revision{Cells} &
    \revision{2} & 
    \revision{$0$} &
    \revision{$0$} &
    \revision{$0$}\\
    \arrayrulecolor{lightgray}
    \hline
    \revision{Ocean Vortices} &
    \revision{2} & 
    \revision{$0$} &
    \revision{$0$} &
    \revision{$0$}\\
    \arrayrulecolor{black}
    \hline
    \revision{Aneurysm} &
    \revision{3} & 
    \revision{$0$} &
    \revision{$1.22\times10^{-07}$} &
    \revision{$8.27\times10^{-04}$}\\
    \arrayrulecolor{lightgray}
    \hline
    \revision{Bonsai} &
    \revision{3} & 
    \revision{$0$} &
    \revision{$4.03\times10^{-07}$} &
    \revision{$3.92\times10^{-03}$}\\
    \arrayrulecolor{lightgray}
    \hline
    \revision{Foot} &
    \revision{3} & 
    \revision{$0$} &
    \revision{$3.28\times10^{-09}$} &
    \revision{$4.19\times10^{-03}$}\\
    \arrayrulecolor{lightgray}
    \hline
    \revision{Neocortical Layer Axon} &
    \revision{3} & 
    \revision{$0$} &
    \revision{$1.14\times10^{-06}$} &
    \revision{$4.38\times10^{-03}$}\\
    \arrayrulecolor{lightgray}
    \hline
    \revision{Dark Sky} &
    \revision{3} & 
    \revision{$0$} &
    \revision{$1.84\times10^{-06}$} &
    \revision{$2.01\times10^{-03}$}\\
    \arrayrulecolor{lightgray}
    \hline
    \revision{Backpack} &
    \revision{3} & 
    \revision{$0$} &
    \revision{$2.85\times10^{-06}$} &
    \revision{$1.25\times10^{-03}$}\\
    \arrayrulecolor{lightgray}
    \hline
    \revision{Head Aneurysm} &
    \revision{3} & 
    \revision{$0$} &
    \revision{$6.24\times10^{-07}$} &
    \revision{$1.20\times10^{-03}$}\\
    \arrayrulecolor{lightgray}
    \hline
    \revision{Chameleon} &
    \revision{3} & 
    \revision{$0$} &
    \revision{$3.14\times10^{-06}$} &
    \revision{$1.43\times10^{-03}$}\\
    \arrayrulecolor{black}
    \hline
    \cellcolor[HTML]{e1e1e1} \revision{Cells} & 
    \cellcolor[HTML]{e1e1e1} \revision{$2$} &
      \cellcolor[HTML]{e1e1e1} \revision{$0$} &
      \cellcolor[HTML]{e1e1e1} \revision{$0$} &
      \cellcolor[HTML]{e1e1e1} \revision{$0$} \\
      \arrayrulecolor{lightgray}
      \hline
      \cellcolor[HTML]{e1e1e1} \revision{Ocean Vortices} & 
      \cellcolor[HTML]{e1e1e1} \revision{$2$} &
        \cellcolor[HTML]{e1e1e1} \revision{$0$} &
        \cellcolor[HTML]{e1e1e1} \revision{$2.99\times10^{-07}$} &
        \cellcolor[HTML]{e1e1e1} \revision{$2.40\times10^{-05}$}\\
       \arrayrulecolor{black}
      \hline
      \cellcolor[HTML]{e1e1e1} \revision{Aneurysm} & 
      \cellcolor[HTML]{e1e1e1} \revision{$3$} &
      \cellcolor[HTML]{e1e1e1} \revision{$0$} &
      \cellcolor[HTML]{e1e1e1} \revision{$5.51\times10^{-05}$} &
      \cellcolor[HTML]{e1e1e1} \revision{$1.27\times10^{-02}$}\\
       \arrayrulecolor{lightgray}
      \hline
      \cellcolor[HTML]{e1e1e1} \revision{Bonsai} & 
      \cellcolor[HTML]{e1e1e1} \revision{$3$} &
        \cellcolor[HTML]{e1e1e1} \revision{$0$} &
        \cellcolor[HTML]{e1e1e1} \revision{$9.42\times10^{-21}$} &
        \cellcolor[HTML]{e1e1e1} \revision{$1.88\times10^{-20}$}\\
       \arrayrulecolor{lightgray}
      \hline
      \cellcolor[HTML]{e1e1e1} \revision{Foot} & 
      \cellcolor[HTML]{e1e1e1} \revision{$3$} &
        \cellcolor[HTML]{e1e1e1} \revision{$0$}&
      \cellcolor[HTML]{e1e1e1} \revision{$5.09\times10^{-21}$}&
      \cellcolor[HTML]{e1e1e1} \revision{$9.17\times10^{-20}$}\\
       \arrayrulecolor{lightgray}
      \hline
      \cellcolor[HTML]{e1e1e1} \revision{Neocortical Layer Axon} & 
      \cellcolor[HTML]{e1e1e1} \revision{$3$} &
    \cellcolor[HTML]{e1e1e1} \revision{$0$} &
      \cellcolor[HTML]{e1e1e1} \revision{$4.22\times10^{-21}$} &
      \cellcolor[HTML]{e1e1e1} \revision{$7.35\times10^{-19}$} \\
      \arrayrulecolor{lightgray}
      \hline
      \cellcolor[HTML]{e1e1e1} \revision{Dark Sky} & 
      \cellcolor[HTML]{e1e1e1} \revision{$3$} &
      \cellcolor[HTML]{e1e1e1} \revision{$0$} &
      \cellcolor[HTML]{e1e1e1} \revision{$1.38\times10^{-04}$} &
      \cellcolor[HTML]{e1e1e1} \revision{$2.25\times10^{-02}$} \\
      \arrayrulecolor{lightgray}
      \hline
      \cellcolor[HTML]{e1e1e1} \revision{Backpack} & 
      \cellcolor[HTML]{e1e1e1} \revision{$3$} &
      \cellcolor[HTML]{e1e1e1} \revision{$0$}&
      \cellcolor[HTML]{e1e1e1} \revision{$8.67\times10^{-22}$}&
      \cellcolor[HTML]{e1e1e1} \revision{$8.41\times10^{-20}$}\\
      \arrayrulecolor{lightgray}
      \hline
      \cellcolor[HTML]{e1e1e1} \revision{Head Aneurysm} & 
      \cellcolor[HTML]{e1e1e1} \revision{$3$} &
      \cellcolor[HTML]{e1e1e1} \revision{$0$} &
      \cellcolor[HTML]{e1e1e1} \revision{$1.06\times10^{-22}$} &
      \cellcolor[HTML]{e1e1e1} \revision{$2.12\times10^{-22}$}\\
      \arrayrulecolor{lightgray}
      \hline
      \cellcolor[HTML]{e1e1e1} \revision{Chameleon} & 
      \cellcolor[HTML]{e1e1e1} \revision{$3$} &
      \cellcolor[HTML]{e1e1e1} \revision{$0$} &
      \cellcolor[HTML]{e1e1e1} \revision{$0$}&
      \cellcolor[HTML]{e1e1e1} \revision{$0$}\\
    \arrayrulecolor{black}
    \hline
    \end{tabular}
    \label{tab_signalPairPreservation}
}
\end{table}

%% file: appendixFullSimplification.tex
\begin{figure*}
    \includegraphics[width=\linewidth]{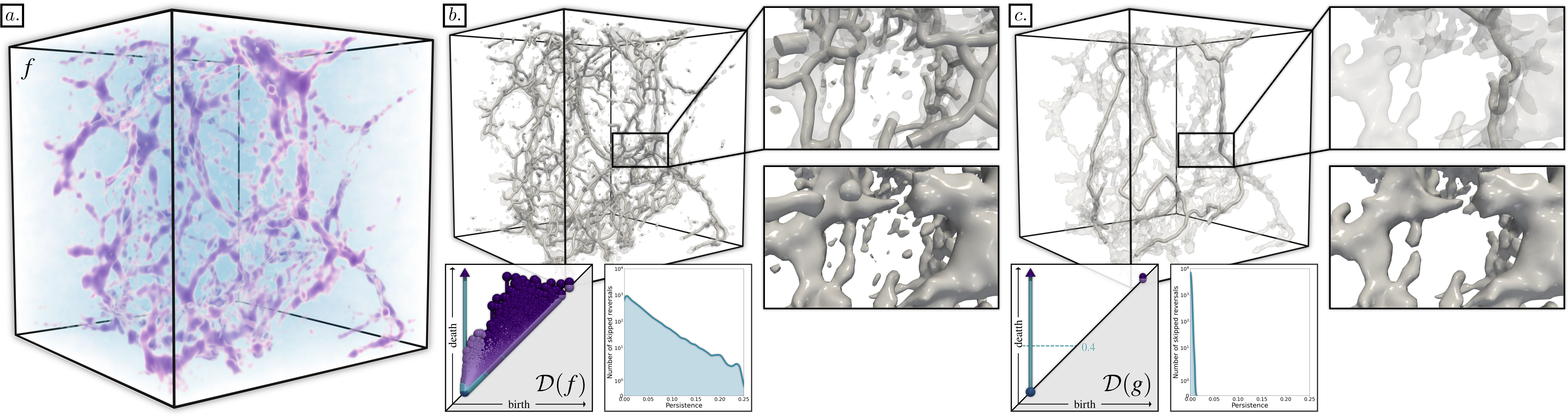}
    \caption{\revision{\emph{Extreme} simplification optimization
    for a
challenging dataset
(\emph{``Dark Sky''}: dark matter
density in a cosmology simulation, $(a)$, \emph{only one signal pair}:
the bar involving the global minimum).
The
geometry of the \emph{cosmic web} \cite{sousbie11, shivashankar2016felix} is
captured $(b)$ by an isosurface (at isovalue $0.4$) and its core filament
structure is extracted by
the
upward
discrete integral lines, started at $2$-saddles above $0.4$.
The latter
structure contains many small-scale loops as many,
persistent saddle connector reversals
could not be performed
(bottom left histogram).
The local minimum $g$  of the simplification energy
found by our solver $(c)$ has a number of
\emph{non-signal} pairs
reduced by $95\%$.
This
results in a
less cluttered visualization,
as the cosmic web has a
drastically simplified
topology (noisy connected components are
removed and most handles are cut, inset zooms). This also induces
fewer skips of persistent saddle connector reversals
(bottom right histogram), here simplifying
all filament
loops
and revealing the core filament structure.}
    }
    \label{fig_darkSky2}
\end{figure*}

\section{\revision{Extreme simplification}}
\revision{Section 5.2 (main manuscript) evaluates our approach from a
qualitative point of view, for various practical scenarios of
simplification:
removing saddle pairs (Figure 1)
or removing  pairs less persistent than an aggressive persistence threshold,
i.e. $0.25$ (Figure 7). In this appendix, we revisit this latter experiment to
stress our approach. Specifically, we consider the challenging \emph{Dark Sky}
dataset (large input diagrams, many saddle pairs, intricate geometry)
and specify
an \emph{extreme}
simplification.
In particular, all the \emph{finite} persistence pairs are considered as
\emph{non-signal} (bars marked with spheres at their extremities,
\autoref{fig_darkSky2})
and only the \emph{infinite} bar involving the global minimum (cropped by
convention at the
globally maximum data value, bar with an upward arrow, \autoref{fig_darkSky2})
is considered as a \emph{signal pair}. The
corresponding results are shown in \autoref{fig_darkSky2}. Specifically, this
figure shows that,
despite this challenging dataset and extreme simplification criterion,
our approach still manages to simplify $95\%$ of the
\emph{non-signal} pairs, which is a slight improvement over the original
experiment reported in the Figure 7 of the main manuscript ($92\%$ for a
persistence threshold of $0.25$). Moreover,
from a qualitative point of view,
all the \emph{filament} loops have been simplified: the
persistence diagram does not contain any \emph{finite} persistence pairs whose
life-span
crosses the death isovalue $0.4$ (dashed horizontal line). Only the
\emph{infinite} bar related to the global minimum (bar with an upward arrow)
crosses it.
In other words, this means that the cosmic web volume (i.e. the sublevel set
for the isovalue
$0.4$) is made of only one connected component and contains no topological
handles.
}